\pdfoutput=1

\documentclass{article}
\usepackage{graphicx}

\usepackage{fancyhdr}

\usepackage{wrapfig}
\usepackage{microtype}
\usepackage{graphicx}
\usepackage{subfigure}
\usepackage{floatrow}
\usepackage{booktabs} 
\usepackage{paralist}
\usepackage{xcolor}
\usepackage{times}
\usepackage{latexsym, caption}
\usepackage[utf8]{inputenc}
\usepackage{times}
\usepackage{latexsym}
\usepackage{algorithm}
\usepackage{algpseudocode}

\usepackage{url}
\usepackage{paralist}
\usepackage{multicol,graphicx,mathrsfs,pifont,amscd,latexsym,url, pifont,epstopdf,setspace,multirow,colortbl, bbm,listings, balance,indentfirst,enumitem,amsmath,amssymb, verbatim,microtype}
\usepackage{graphicx,booktabs, paralist}
\usepackage{blindtext}
\usepackage[nocompress]{cite}

\usepackage{color}
\definecolor{darkblue}{rgb}{0.0, 0.2, 0.6}
\usepackage[colorlinks,
linkcolor=black,
anchorcolor=darkblue,
citecolor=darkblue,
urlcolor=darkblue
]{hyperref}

\usepackage[final, nonatbib]{neurips_2023}

\algnewcommand\algorithmicswitch{\textbf{switch}}
\algnewcommand\algorithmiccase{\textbf{case}}
\algnewcommand\algorithmicassert{\texttt{assert}}
\algnewcommand\Assert[1]{\State \algorithmicassert(#1)}%
\algdef{SE}[SWITCH]{Switch}{EndSwitch}[1]{\algorithmicswitch\ #1\ \algorithmicdo}{\algorithmicend\ \algorithmicswitch}%
\algdef{SE}[CASE]{Case}{EndCase}[1]{\algorithmiccase\ #1}{\algorithmicend\ \algorithmiccase}%
\algtext*{EndSwitch}%
\algtext*{EndCase}%

\usepackage[utf8]{inputenc} 
\usepackage[T1]{fontenc}    
\usepackage{url}            
\usepackage{booktabs}       
\usepackage{amsfonts}       
\usepackage{nicefrac}       
\usepackage{microtype}      
\usepackage{xcolor}         
\usepackage{amsmath}

\newcommand{\framework}{AVIS\xspace}

\title{\framework: Autonomous Visual Information Seeking\\ with Large Language Model Agent}

\author{%
  Ziniu Hu$^{12}$\thanks{This work was done when Ziniu was an intern at Google.}\\
  \And
  Ahmet Iscen$^2$\\
  \And
  Chen Sun$^2$\\
  \And
  Kai-Wei Chang$^1$\\
  \And
  Yizhou Sun$^1$\\ 
  \And
  David A Ross$^2$\\
  \And
  Cordelia Schmid$^2$\\
  \And
  Alireza Fathi$^2$\\  \And
   $^1$\textmd{University of California, Los Angeles},\  \ $^2$\textmd{Google Research}
}

\begin{document}


\def\eg{\emph{e.g }\onedot}
\def\ie{\emph{i.e }\onedot} 
\newcommand{\head}[1]{{\noindent\bf #1}}  

\maketitle

\begin{abstract}
  In this paper, we propose an autonomous information seeking visual question answering framework, \framework.
  Our method leverages a Large Language Model (LLM) to dynamically strategize the utilization of external tools and to investigate their outputs via tree search, thereby acquiring the indispensable knowledge needed to provide answers to the posed questions.
  Responding to visual questions that necessitate external knowledge, such as "What event is commemorated by the building depicted in this image?", is a complex task. This task presents a combinatorial search space that demands a sequence of actions, including invoking APIs, analyzing their responses, and making informed decisions.
  We conduct a user study to collect a variety of instances of human decision-making when faced with this task. This data is then used to design a system comprised of three components: an LLM-powered planner that dynamically determines which tool to use next, an LLM-powered reasoner that analyzes and extracts key information from the tool outputs, and a working memory component that retains the acquired information throughout the process.
  The collected user behavior serves as a guide for our system in two key ways. First, we create a transition graph by analyzing the sequence of decisions made by users. This graph delineates distinct states and confines the set of actions available at each state. 
  Second, we use examples of user decision-making to provide our LLM-powered planner and reasoner with relevant contextual instances, enhancing their capacity to make informed decisions.
  We show that AVIS achieves state-of-the-art results on knowledge-intensive visual question answering benchmarks such as Infoseek~\cite{chen2023pretrained} and OK-VQA~\cite{DBLP:conf/cvpr/MarinoRFM19}.

\end{abstract}

\section{Introduction}

Large language models (LLMs), such as GPT3~\cite{DBLP:conf/nips/BrownMRSKDNSSAA20}, LaMDA~\cite{adiwardana2020humanlike}, PALM~\cite{DBLP:journals/corr/abs-2204-02311}, BLOOM~\cite{DBLP:journals/corr/abs-2211-05100} and LLaMA~\cite{touvron2023llama}, have showcased the capacity to memorize and utilize a significant amount of world knowledge. They demonstrate emerging abilities~\cite{wei2022emergent} like in-context learning~\cite{DBLP:conf/nips/BrownMRSKDNSSAA20}, code generation~\cite{DBLP:journals/corr/abs-2203-07814}, and common sense reasoning~\cite{DBLP:conf/emnlp/MadaanZ0YN22}. Recently, there is a growing focus towards adapting LLMs to handle multi-modal inputs and outputs involving both vision and language. Noteworthy examples of such visual language models (VLMs) include GPT4~\cite{openai2023gpt4}, Flamingo~\cite{DBLP:journals/corr/abs-2204-14198} and PALI~\cite{DBLP:journals/corr/abs-2209-06794}. They set the state of the art for several tasks, including image captioning, visual question answering, and open vocabulary recognition. 

While LLMs excel beyond human capabilities in tasks involving textual information retrieval, the current state of the art VLMs perform inadequately on datasets designed for visual information seeking such as Infoseek~\cite{chen2023pretrained} and OK-VQA~\cite{DBLP:conf/cvpr/MarinoRFM19}. Many of the visual questions in these datasets are designed in such a way that they pose a challenge even for humans, often requiring the assistance of various APIs and web search to obtain the answer. Examples of such questions include "where is this church located?", "what species of butterfly is this?", or "what is the brand of this dress?". 

Current state-of-the-art vision-language models (VLMs) find it challenging to answer such questions for several reasons. Firstly, they are not trained with objectives that encourage them to discern fine-grained categories and details within images. Secondly, they utilize a relatively smaller language model compared to state-of-the-art Large Language Models (LLMs), which constrains their reasoning capabilities. Lastly, they do not compare the query image against a substantial corpus of images associated with varying metadata, unlike systems that employ image search techniques.


\begin{figure}[t]
    \includegraphics[bb=0 0 1280 1280, width=0.822\columnwidth]{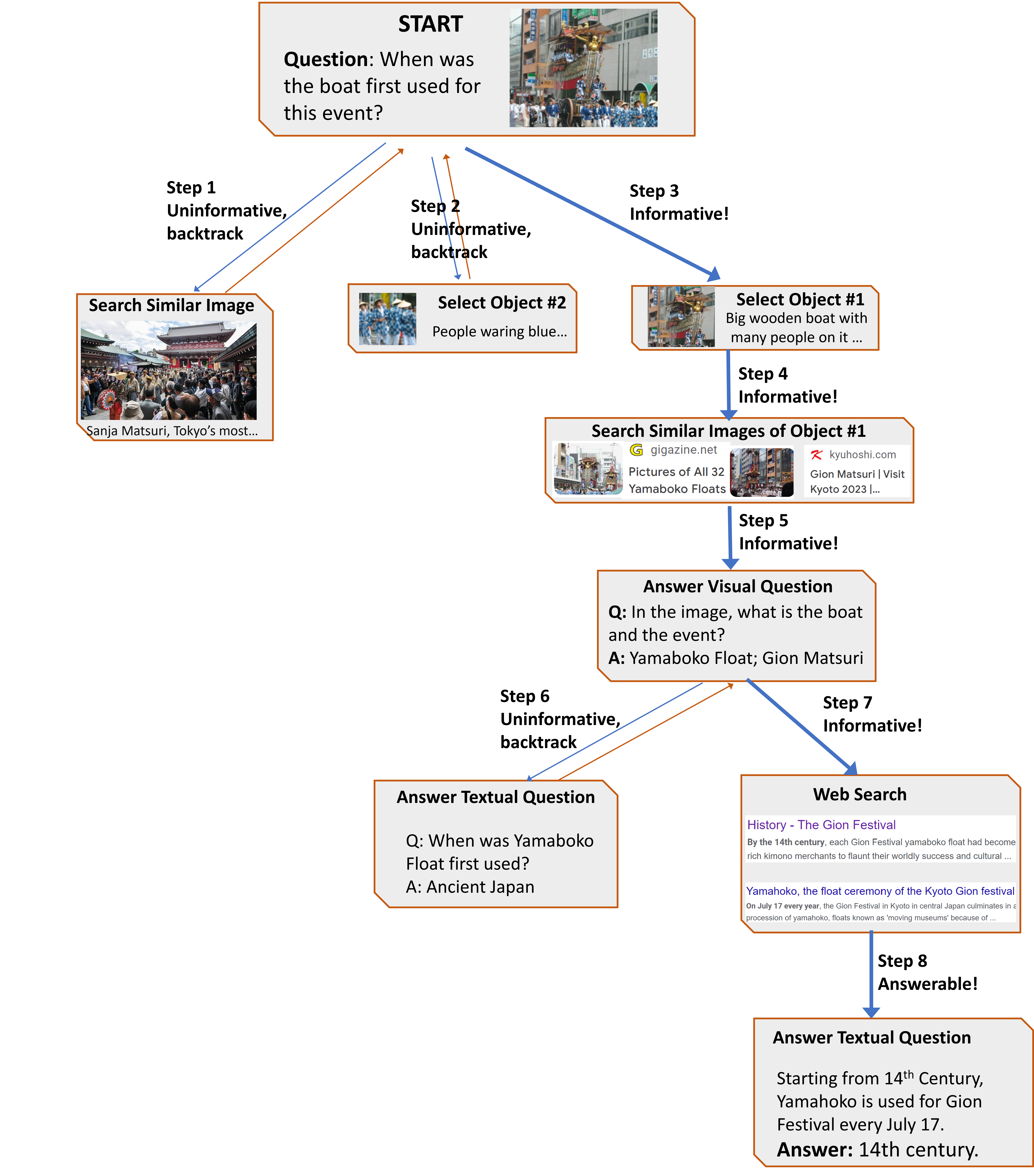}
    \vspace*{-.1in}
    \caption{An example of \framework's generated workflow for answering a challenging visual question using LLM with tree search to use tools. The input image is taken from the Infoseek dataset.}
    \label{fig:tease}
\end{figure}



To overcome these challenges, we introduce a novel method in this paper that achieves state-of-the-art results on visual information seeking tasks by enabling a \textbf{LLM Agent use tools via tree-search decision-making}. We use three types of tools: (i) computer vision tools such as object detection, OCR, image captioning models, and VQA models, which aid in extracting visual information from the image, (ii) a web search tool that assists in retrieving open world knowledge and facts, and (iii) an image search tool that enables us to glean relevant information from metadata associated with visually similar images. Our approach utilizes an LLM-powered planner to dynamically determine which tool to use at each step and what query to send to it. Furthermore, we employ an LLM-powered reasoner that scrutinizes the output returned by the tools and extracts the crucial information from them. To retain the information throughout the process, we use a working memory component. Figure~\ref{fig:tease} shows an example information seeking process performed by our method. 

Several recent studies~\cite{wu2023visual, yang2023mmreact, gupta2022visual, surís2023vipergpt, lu2023chameleon} have enhanced LLMs with APIs to handle multi-modal vision-language inputs. These systems generally employ a two-stage strategy, namely \textit{plan} and \textit{execute}. Initially, the LLM breaks down a question into a plan, typically represented as a structured program or a sequence of instructions. Following this, the necessary APIs are activated to collect the required information. While this method has shown potential in elementary visual-language tasks, it frequently fails in more complex real-world situations. In such cases, a comprehensive plan cannot be inferred merely from the initial question. Instead, it necessitates dynamic modifications based on real-time feedback.

The primary innovation in our proposed method lies in its dynamic decision-making capability. Answering visual information seeking questions is a highly complex task, requiring the planner to take multiple steps. At each of these steps, the planner must determine which API to call and what query to send. It is unable to predict the output of complex APIs, such as image search, or to anticipate the usefulness of their responses prior to calling them. Therefore, unlike previous methods that pre-plan the steps and API calls at the beginning of the process, we opt for a dynamic approach. We make decisions at each step based on the information acquired from previous API calls, enhancing the adaptability and effectiveness of our method. 

We conduct a user study to gather a wide range of instances of human decision-making when using APIs to answer questions related to visual information seeking. From this data, we formulate a structured framework that directs the Large Language Model (LLM) to use these examples for making informed decisions regarding API selection and query formulation.
The collected user behavior informs our system in two significant ways. First, by analyzing the sequence of user decisions, we construct a transition graph. This graph delineates distinct states and constrains the set of actions available at each state. Second, we use the examples of user decision-making to guide our planner and reasoner with pertinent contextual instances. These contextual examples contribute to improving the performance and effectiveness of our system.


The primary contributions of this paper can be summarized as follows:
\begin{compactitem}
\item We propose a novel visual question answering framework that leverages a large language model (LLM) to dynamically strategize the utilization of external tools and to investigate their outputs, thereby acquiring the necessary knowledge needed to provide answers to the posed questions.
\item We leverage the human decision-making data collected from a user study to develop a structured framework. This framework guides the Large Language Model (LLM) to utilize examples of human decision-making in making informed choices concerning API selection and query construction.
\item Our method achieves state-of-the-art results on knowledge-based visual question answering benchmarks such as Infoseek~\cite{chen2023pretrained} and OK-VQA~\cite{DBLP:conf/cvpr/MarinoRFM19}. Notably, We achieve an accuracy of $50.7\%$ on the Infoseek (unseen entity split) dataset which is significantly higher than the results achieved by PALI~\cite{DBLP:journals/corr/abs-2209-06794} with accuracy of $16.0\%$.

\end{compactitem}

\section{Related Work}

\paragraph{Augmenting LLMs with Tools.} Large Language Models(LLMs) have shown impressive language understanding~\cite{radford2018improving}, and even reasoning capabilities~\cite{DBLP:conf/nips/Wei0SBIXCLZ22}.
Nevertheless, certain limitations of LLMs are evident, due to their intrinsic characteristics. 
Such limitations include providing up-to-date answers based on external knowledge or performing mathematical reasoning. 
Consequently, a recent surge of techniques have integrated LLMs with various external tools~\cite{mialon2023augmented}. 
For example, TALM~\cite{parisi2022talm} and ToolFormer~\cite{schick2023toolformer} use in-context learning to teach the language model how to better leverage various tools on benchmarks such as question answering and mathematical reasoning. 

In the computer vision domain, LLMs also show significant improvements when combined with external visual tools.
For example, Visual ChatGPT~\cite{wu2023visual} and MM-ReAct~\cite{yang2023mmreact} enable LLMs to call various vision foundation models as tools to understand visual inputs, and even better control the image generation. 
VisProg~\cite{gupta2022visual} and ViperGPT~\cite{surís2023vipergpt} explore the decomposition of visual language tasks into programs, where each line corresponds to general code or a visual API.
Chameleon~\cite{lu2023chameleon} uses an LLM as a natural language planner to infer the appropriate sequence of tools to utilize, and then executes these tools to generate the final response.

Most of these previous works follow a plan-then-execute paradigm, i.e., i) they pre-plan the sequence of actions (API calls) that they will take (either hard coded or using code generation); and ii) they execute the generated plan. 
One drawback of such an approach is that it cannot update and improve its plan based on the output of the tools it calls. 
This is not a trivial problem, as it requires to predict the output quality of each tools beforehand.
In contrast, our proposed method allows the system to dynamically decide its next steps based on the output it receives from the tools at each step.

\paragraph{Decision Making with LLM as an Agent.}
There has also been a surge of interest in applying Large Language Models (LLMs) as autonomous agents.
These agents are capable of interacting with external environments, making dynamic decisions based on real-time feedback, and consequently achieving specific goals. 
For example, WebGPT~\cite{DBLP:journals/corr/abs-2112-09332} enables an LLM to access real-time information from the web search engines. 
ReAct~\cite{DBLP:journals/corr/abs-2210-03629} further improves external search engine usage via the self-reasoning of LLM in an interleaved manner.
Similar ideas have also been adopted for robotic action planning. 
SayCan~\cite{DBLP:conf/corl/IchterBCFHHHIIJ22}, for instance, uses LLMs to directly predict robot actions, and PALM-E~\cite{DBLP:journals/corr/abs-2303-03378} further fine-tunes LLMs to make better decisions based on instructions and open web media.

When compared to works that follow a plan-then-execute paradigm, these AI agents exhibit increased flexibility, adjusting their actions based on the feedback that they receive. 
However, many of these methods do not restrict the potential tools that can be invoked at each stage, leading to an immense search space. 
This becomes particularly critical for web search APIs~\cite{lens, google} that return extensive result lists and span a combinatorial search space of multiple tools. 
Consequently, even the most advanced LLMs today can fall into infinite loops or propagate errors.
To alleviate this issue, we propose restricting and guiding LLMs to mimic human behavior when solving complex visual questions with APIs. 
This idea is similar to the AI alignment research~\cite{DBLP:conf/nips/Ouyang0JAWMZASR22, liu2023visual} that teaches LLMs to follow human instructions. 
The difference is that our model only uses the human prior at the decision-making stage via prompt guidance, instead of re-training the model.

One concurrent work Tree-Of-Thought (ToT)~\cite{DBLP:journals/corr/abs-2305-10601} also utilize tree search guided by a self-critic reward model to find optimal path of problem solving. Compared with this concurrent work, our AVIS further constrains the tree search via a human-defined transition graph, and guide the decision-making via a dynamic prompt manager. In addition, though AVIS is designed for tool-use, the success of ToT shows that such idea can be generally improve many LLM Reasoning tasks.



\section{Method}

\subsection{General Framework}

Our approach employs a dynamic decision-making strategy designed to respond to visual information-seeking queries. Our system is comprised of three primary components. First, we have a planner $\mathcal{P}$, whose responsibility is to determine the subsequent action, including the appropriate API call and the query it needs to process. Second, we have a working memory $\mathcal{M}$ that retains information about the results obtained from API executions. Lastly, we have a reasoner $\mathcal{R}$, whose role is to process the outputs from the API calls. It determines whether the obtained information is sufficient to produce the final response, or if additional data retrieval is required.

\begin{minipage}{.55\textwidth}
  \centering
  \includegraphics[bb=0 0 650 500,width=0.98\textwidth]{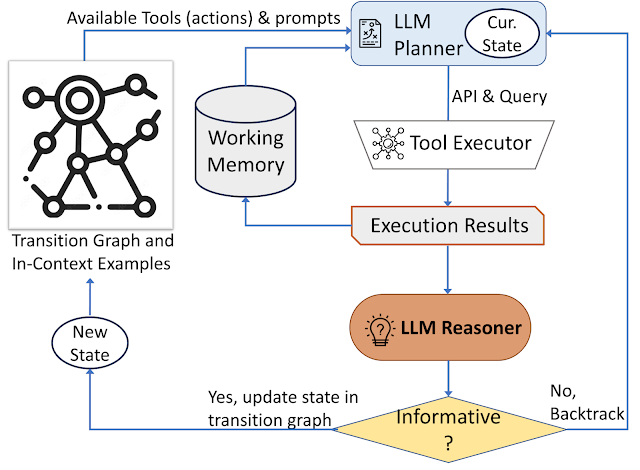}
  \vspace*{-.1in}
  \captionof{figure}{AVIS employs dynamic decision-making to \textbf{plan} (find optimal tool and query), execute results, and then \textbf{reason} (estimate whether continue or backtrack).} \label{fig:graph}
\end{minipage}
\begin{minipage}{.42\textwidth}
\vspace{-7mm}
\begin{algorithm}[H]
\caption{Planner $\mathcal{P}(state,\mathcal{G},\mathcal{E},\mathcal{M})$}\label{alg:planner}
\begin{algorithmic}[1]
\State $\mathcal{A}_s \gets \phi(state,\mathcal{G},\mathcal{M})$ \Comment{Get the list of feasible actions $\mathcal{A}_s$ given the current state from transition graph and the information in the working memory}
\State $\mathcal{E}_s \gets \theta(\mathcal{E},\mathcal{A}_s)$ \Comment{Get a list of in-context examples related to actions $\mathcal{A}_s$}
\State $p_s \gets \psi(\mathcal{E}_s,\mathcal{M})$ \Comment{Build a prompt based on the in-context examples $\mathcal{E}_s$ and the current working memory $\mathcal{M}$}
\State $t_s,q_s \gets LLM(p_s)$ \Comment{Decide the next tool $t_s$ to use and the query $q_s$ to pass by feeding the prompt $p_s$ to LLM}
\end{algorithmic}
\end{algorithm}
\end{minipage}%

\begin{minipage}{.5\textwidth}
  \centering
  \includegraphics[width=0.75\textwidth]{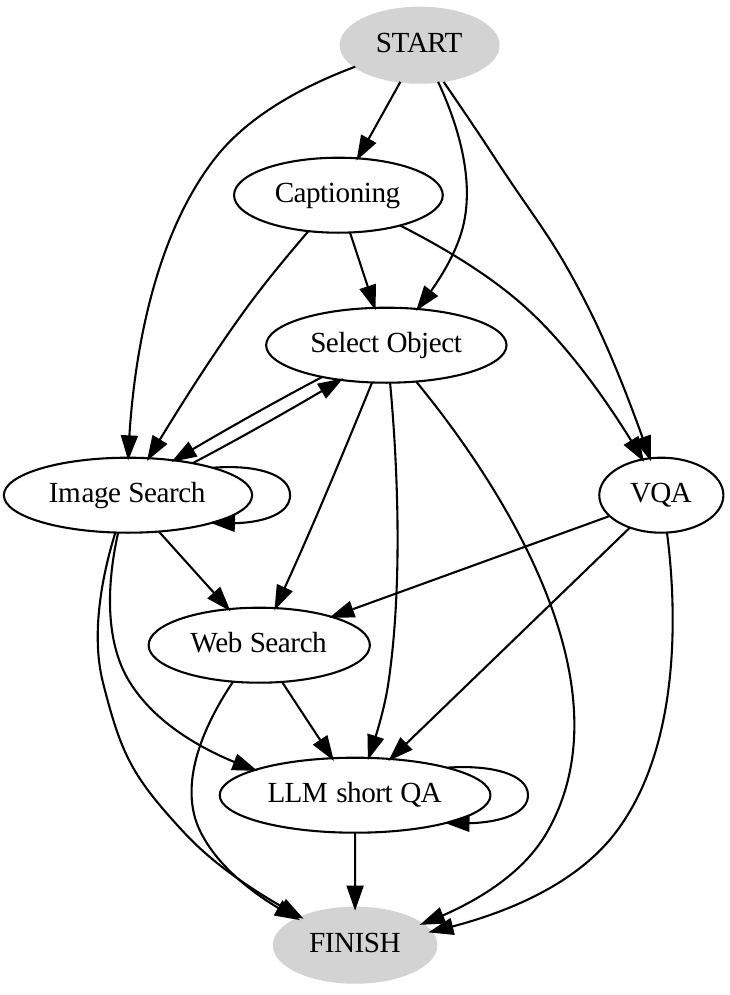}
  \vspace*{-.1in}
  \captionof{figure}{Transition graph $\mathcal{G}$ defines feasible actions the planner can take. This graph is induced by our user study introduced in Sec.~\ref{sec:study}.} \label{fig:graph}
\end{minipage}
\hspace{1mm}
\begin{minipage}{.5\textwidth}
\vspace{-5mm}
\begin{algorithm}[H]
\caption{\framework Decision Making Workflow}\label{alg:cap}
\begin{algorithmic}[1]
\State $\mathcal{M} \gets \{input\}$, $state \gets \texttt{START}$
\State $t_s,q_s \gets \mathcal{P}(state,\mathcal{G},\mathcal{E},\mathcal{M})$   \Comment{Call the planner $\mathcal{P}$ to decide the next tool to use $t_s$ and the query to pass to it $q_s$ } 
\State $o_s \gets \texttt{Exec}(t_s, q_s)$ \Comment{Call tool $t_s$ with query $q_s$ and get output $o_s$}
\State $\hat{o}_s \gets \mathcal{R}(o_s, \mathcal{M})$ \Comment{Process the output and extract the key info $\hat{o_s}$ using the reasoner $\mathcal{R}$}
\State $\mathcal{M}.\texttt{add}(\hat{o}_s)$ \Comment{Update the working memory }
\Switch{$\hat{o}_s$}
    \Case{$\hat{o}_s$ is not informative}
        \State $\texttt{goto}(2)$ \Comment{Go to line 2 to make decision at the same state, excluding $t_s$.}
    \EndCase
    \Case{$\hat{o}_s$ has useful information}
        \State $state \gets t_s$ \Comment{Update state}
        \State $\texttt{goto}(2)$ \Comment{Go to line 2 to make decision for the next state.}
    \EndCase
    \Case{$\hat{o}_s$ is ready as final answer}
        \State $ans \gets \hat{o}_s$ \Comment{Output answer}
    \EndCase
\EndSwitch
\end{algorithmic}
\end{algorithm}
\end{minipage}%

Considering the potential intricacy of the task, we conduct a user study to gather a broad range of examples of human decision-making process, when using tools to respond to visual information-seeking queries (we introduce the details of data collection in Sec.~\ref{sec:study}). This helps us to establish a structured framework for decision-making. We utilize the data collected from this study to construct a transition graph $\mathcal{G}$ shown in Figure~\ref{fig:graph}, which outlines all the possible actions at each given state. Additionally, we employ real-life decision-making examples $\mathcal{E}$, i.e., users choose which tool at  different states,
to guide the planner in choosing the appropriate action at each stage of the process.

The Algorithm~\ref{alg:planner} presents the operations of the planner $\mathcal{P}$. 
The planner undertakes a series of steps each time a decision is required regarding which tool to employ and what query to send to it. 
Firstly, based on the present $state$, the planner provides a range of potential subsequent actions $\mathcal{A}_s$.
The potential action space $\mathcal{A}_s$ may be large, making the search space intractable.
To address this issue, the planner refers to the human decisions from the transition graph $\mathcal{G}$ to eliminate irrelevant actions.
The planner also excludes the actions that have already been taken before and are stored in the working memory $\mathcal{M}$. Formally, this procedure is $\mathcal{A}_s \gets \phi(state,\mathcal{G},\mathcal{M})$.

Next, it collects a set of relevant in-context examples $\mathcal{E}_s$ that are assembled from the decisions previously made by humans during the user study relevant to actions $\mathcal{A}_s$, that is $\mathcal{E}_s \gets \theta(\mathcal{E},\mathcal{A}_s)$. 
With the gathered in-context examples $\mathcal{E}_s$ and the working memory $\mathcal{M}$ that holds data collected from past tool interactions, the planner formulates a prompt, denoted by $p_s \gets \psi(\mathcal{E}_s,\mathcal{M})$. 
The prompt $p_s$ is then sent to the LLM which returns a structured answer, determining the next tool $t_s$ to be activated and the query $q_s$ to be dispatched to it.
We denote this action by $t_s,q_s \gets LLM(p_s)$. 
This design allows the planner to be invoked multiple times throughout the process, thereby facilitating dynamic decision-making that gradually leads to answering the input query.


The Algorithm~\ref{alg:cap} shows the overall decision-making workflow of \framework. The entire process repeats until a satisfactory answer is produced. Initially, the working memory is populated only with the input visual question $I$, and the initial $state$ is set to $\texttt{START}$. 
At each iteration, we first invoke the planner $\mathcal{P}$ to determine the next tool and the query to employ, as outlined in Algorithm~\ref{alg:planner}. Subsequently, the selected external tool executes and delivers its output $o_s$. The output from the tools can be quite diverse, ranging from a list of identified objects, to a collection of similar images with their captions, to snippets of search results or knowledge graph entities.


Therefore, we employ a reasoner $\mathcal{R}$ to analyze the output $o_s$, extract the useful information  and decide into which category the tool output falls: informative, uninformative, or final answer. Our method utilizes the LLM with appropriate prompting and in-context examples to perform the reasoning. If the reasoner concludes that it's ready to provide an answer, it will output the final response, thus concluding the task. If it determines that the tool output is uninformative, it will revert back to the planner to select another action based on the current state. If it finds the tool output to be useful, it will modify the state and transfer control back to the planner to make a new decision at the new state. 

Our approach, which employs dynamic decision-making coupled with backtracking, differs from previous methods~\cite{surís2023vipergpt, lu2023chameleon} that follow a plan-then-execute paradigm. Our system is structured to make decisions grounded to the results of current executions and to conduct iterative searches for tool combinations. This process eventually yields the most effective strategy to accomplish the task.

\subsection{Tools and their APIs}



To respond effectively to visual queries that necessitate in-depth information retrieval, it's important to equip \framework with a comprehensive suite of tools. In this section, we describe these tools.

\head{Image Captioning Model}: We employ the PALI 17B~\cite{DBLP:journals/corr/abs-2206-14286} captioning model, which obtains state-of-the-art results for image captioning. This tool has the capability to generate captions for either the entire image or for a cropped image corresponding to the bounding box of a detected object.

\head{Visual Question Answering Model}: We utilize the PALI 17B~\cite{DBLP:journals/corr/abs-2206-14286} VQA model, which has been fine-tuned on the VQA-v2~\cite{DBLP:journals/ijcv/GoyalKASBP19} dataset. This tool takes an image and a question as inputs and provides a text-based answer as the output.

\head{Object Detection}: We use an object detector trained on a super-set of Open Images dataset~\cite{OpenImages} categories that is provided by Google Lens API~\cite{lens}. We use high confidence threshold to only keep the top-ranked detected boxes for the input image.

\head{Image Search}: We utilize Google Image Search to obtain a broad range of information related to the image crop of a detected box as provided in Google Lens API~\cite{lens}. This information encompasses various details, such as knowledge graph entities, titles of associated products, and captions of analogous or identical images. The availability of these details can vary based on the image crop input provided to Google Image Search. When it comes to decision-making, our planner considers the utilization of each piece of information as a separate action. This is due to the fact that each information could contain hundreds of tokens that necessitate complex processing and reasoning.

\head{OCR}: In some cases, images may include textual content such as street names or logos. To detect and utilize this text, we take advantage of the Optical Character Recognition (OCR) feature available in the Google Lens API~\cite{lens}.



\head{Web Search}: Web search enables our approach to acquire up-to-date world knowledge and retrieve relevant documents on any topic of interest. For this objective, we employ the Google Web Search API~\cite{google}. It accepts a text-based query as input and produces the following outputs: (i) related document links and snippets, (ii) in certain instances, a knowledge panel providing a direct answer to the query, and (iii) up to five questions that are related to the input query. If a knowledge panel is available, we parse it into a sentence or a few sentences that summarize its information.

\head{LLM short QA}: We incorporate a Language Model (LLM) powered question-answering component as another tool. 
This tool accepts a query in text form and produces an answer also in text form. It is important to note that the use of the LLM here as a question-answering tool is distinct from its role in the planner or reasoner as outlined in Alg.~\ref{alg:planner} and Alg.~\ref{alg:cap}.

\begin{figure}[t]
\vspace*{-.4in}
    \centering
    \includegraphics[width=0.95\columnwidth]{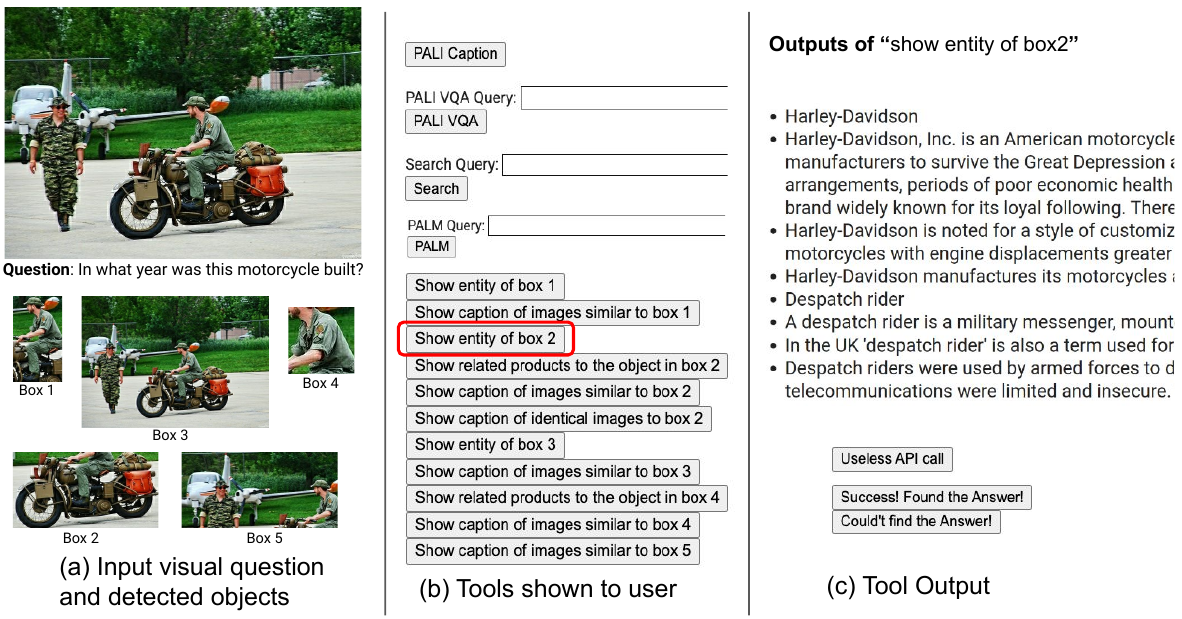}
    \vspace*{-.1in}
    \caption{We conduct a user study to gather examples of user decision-making when responding to visual information-seeking questions. Given a visual question as depicted in (a), the user makes a series of tool calls using the available APIs shown in (b). Each tool call yields an output which the user reviews whether it is useful and determines the subsequent action, illustrated in (c).}
    \label{fig:ui}
\end{figure}

\subsection{Gathering User Behavior to Inform LLM Decision Making}\label{sec:study}
Many of the visual questions in datasets such as Infoseek~\cite{chen2023pretrained}, Oven~\cite{hu2023opendomain} and OK-VQA~\cite{DBLP:conf/cvpr/MarinoRFM19} ask for fine-grained answers, which poses a challenge even for humans, often requiring the assistance of various APIs and web searches for answers. Figure~\ref{fig:ui}(a) illustrates an example visual question taken from the OK-VQA~\cite{DBLP:conf/cvpr/MarinoRFM19} dataset. 
In order to gather insights into human decision-making process, we carried out a user study. More specifically, our goal is to understand how humans utilize external tools to answer visual queries that involve seeking information.

The user is equipped with an identical set of tools as our method. 
They are presented with the input image and question, along with image crops for each detected object. 
Additionally, tools like PALI Caption, PALI VQA, PALM, and Web Search are made available to the user. 
Furthermore, based on the information obtained through image search for each cropped image, the user is offered one or multiple buttons associated with each box. These buttons provide the user with the ability to access diverse information pertaining to the image crop of the box. This includes details such as corresponding knowledge graph entities, captions of similar images, titles of associated related products, and captions of identical images. An example set of tools and APIs are shown in Figure~\ref{fig:ui}(b).

When the user initiates an action, such as clicking on a button or submitting a query to web search, PALM, or PALI VQA, the corresponding tool is invoked, and the resulting output is displayed to the user. We record the sequence of actions taken by the user and the outputs that they receive at each step. 
For instance, in Figure~\ref{fig:ui}, we show an example of how a user needs to perform four actions to answer the question: \emph{i)}~display entities in box 2, \emph{ii)} show the caption of similar images to box 2, \emph{iii)} conduct a search for \emph{"In what year was Harley-Davidson XA built?"}, and \emph{iv)} utilize PALM using the combination of the search output and the question \emph{"In what year was Harley-Davidson XA built?"}. 
When the user is prepared to proceed to the next question, they click on either of the two buttons: "Success! Found the Answer!" or "Couldn't Find the Answer." Subsequently, a new visual question is presented to them. 

The collected user behavior serves as a guide for our system in two key ways. Firstly, we construct a transition graph by analyzing the sequence of decisions made by users. This graph defines distinct states and restricts the available set of actions at each state. For example, at the START state, the system can take only one of these three actions: PALI caption, PALI VQA, or object detection. Figure~\ref{fig:graph} illustrates the transition graph that has been constructed based on the decision-making process of the users. Secondly, we utilize the examples of user decision-making to guide our planner and reasoner with relevant contextual instances. These in-context examples aid in enhancing the performance and effectiveness of our system. 

We conducted a user study involving 10 participants who collectively answered a total of 644 visual questions. During the study, we presented users with visual questions that were randomly selected from both the Infoseek~\cite{chen2023pretrained} and OK-VQA~\cite{DBLP:conf/cvpr/MarinoRFM19} datasets. This approach allowed us to provide the participants with a varied and diverse set of visual questions to assess and respond to. We show the details for this study as well as example prompts in the Appendix.


\section{Experiments}
We evaluate \framework on two visual question answering datasets: \emph{i)} OK-VQA~\cite{DBLP:conf/cvpr/MarinoRFM19}, which requires common-sense knowledge not observed in given image; and \emph{ii)} Infoseek$_{\text{wikidata}}$~\cite{chen2023pretrained}, which further necessitates more fine-grained information that cannot be covered by common sense knowledge. 

\textbf{Experimental Setup.}
We follow the decision-making workflow in Alg.~\ref{alg:cap} to implement \framework to solve visual questions. For the Planner, we write the basic instructions for describing each tool, and keep a pool of real user behavior when they select each tool, which we collected in the user study. At each step $s$, we prepare the prompt based on the feasible action lists $\mathcal{A}_s$. 
For the Reasoner, we write the prompt for all APIs that return a long list of results, including \textit{Object Detection}, \textit{Product Detection}, \textit{Web Image Search} and \textit{Web Text Search}, that guides reasoner to extract the relevant information.
Note that we design the reasoner in a way such that the ``uninformative'' answers can be detected.
In order to support this, we manually prepare several bad examples that do not provide any useful information, pass it to the reasoner as a part of the prompt. We show the detailed prompts for these two modules in the Appendix.

We use the frozen PALM 540B language model~\cite{DBLP:journals/corr/abs-2204-02311} for both the planner and the reasoner, with deterministic generation ensured by setting the temperature parameter to zero. 
We use 10 examples as in-context prompts for each dataset, and report the VQA accuracy~\cite{DBLP:journals/ijcv/GoyalKASBP19} as the evaluation metric.

\begin{table}[t]
\centering
\begin{tabular}{lcc}
\toprule  
\textbf{Model}  & Unseen Entity & Unseen Question \\             
\midrule
PALM~\cite{DBLP:journals/corr/abs-2204-02311} (Q-only, few-shot)                   & 3.7     & 5,1                           \\
OFA~\cite{DBLP:journals/corr/abs-2206-08916} (fine-tune)                           & 9.7    &   14.8                        \\
PALI~\cite{DBLP:journals/corr/abs-2209-06794} (VQA, zero-shot)          & 1.8  & 2.2    \\
PALI~\cite{DBLP:journals/corr/abs-2209-06794} (fine-tune)                      & 16.0    &    \underline{20.7}                      \\ 
PALM~\cite{DBLP:journals/corr/abs-2204-02311} w/ CLIP~\cite{DBLP:conf/icml/RadfordKHRGASAM21} (few-shot + external knowledge)                  & \underline{21.9}    &    18.6                        \\ 
FiD~\cite{DBLP:conf/acl/Yu0F0WXRY022} w/ CLIP~\cite{DBLP:conf/icml/RadfordKHRGASAM21} (fine-tune + external knowledge)                  & 20.7    &    18.1                        \\ 
\midrule
\multicolumn{3}{c}{(---baselines without dynamic decision making, sequentially execute the tools---)} \\
baseline-PALM w/ (PALI$^*$, few-shot)              & 12.8    &   14.9                       \\
baseline-PALM w/ (PALI$^*$ + Object, few-shot)      & 31.3  &  36.1    \\
baseline-PALM w/ (PALI$^*$ + Object + Search, few-shot) &  36.1   &   38.2                      \\ \midrule
\textbf{\framework} (ours, few-shot)    & \textbf{50.7}   &   \textbf{56.4}\\ 
~~ w/o PALI$^*$    & 47.9 & 54.2 \\
~~ w/o Object      & 41.2 & 48.4 \\
~~ w/o Search       & 42.5  & 49.6\\
\bottomrule
\end{tabular}
\caption{\textbf{Visual Question Answering} results (accuracy) on Infoseek$_{\text{Wikidata}}$. The first four rows are results from their paper that do not use external knowledge, and the next two are from their paper that use CLIP as knowledge source. The tool PALI$^*$ denotes the frozen multi-task PALI-17B model for both visual question answering and image captioning. Object means object detection, and search means image and text search.}
\label{table:infoseek}
\end{table}

\begin{table}[t!]
\centering
\begin{tabular}{llc} \toprule
& \textbf{Model}  & \textbf{Accuracy ($\%$)}\\ \midrule
\multirow{5}{*}{\rotatebox[origin=c]{90}{{Supervised }}} & KRISP~\cite{DBLP:conf/cvpr/MarinoCP0R21}   &  38.4 \\
~&KAT~\cite{DBLP:conf/naacl/GuiWH0BG22}    &  54.4   \\
~&ReVIVE~\cite{DBLP:journals/corr/abs-2206-01201}   & 58.0 \\
~&REVEAL~\cite{DBLP:journals/corr/abs-2212-05221}  & 59.1 \\ 
~&PALI~\cite{DBLP:journals/corr/abs-2209-06794} (OK-VQA, finetune)& \underline{64.5} \\ \midrule
\multirow{4}{*}{\rotatebox[origin=c]{90}{{Zero-shot}}}
~&PALI~\cite{DBLP:journals/corr/abs-2209-06794} (VQA, zero-shot)  & 41.6 \\
~&PICa-Full~\cite{DBLP:journals/corr/abs-2109-05014}   & 48.0    \\
~& Flamingo (zero-shot)~\cite{DBLP:journals/corr/abs-2204-14198}  & 50.6 \\
~& BLIP-2~\cite{DBLP:journals/corr/abs-2301-12597} & 45.9 \\ \midrule
\multirow{11}{*}{\rotatebox[origin=c]{90}{{Few-shot}}}
~& ViperGPT (one-shot)~\cite{surís2023vipergpt} & 51.9 \\
~& Flamingo (few-shot)~\cite{DBLP:journals/corr/abs-2204-14198}  & 57.8 \\ 
\cmidrule{2-3} 
~ & (baselines without dynamic decision making, sequentially executing the tools) & ~ \\ 
~&baseline-PALM w/ (PALI$^*$)  & 44.3\\
~&baseline-PALM w/ (PALI$^*$+Object)  & 38.2 \\
~&baseline-PALM w/ (PALI$^*$+Object + Search)  & 47.9 \\ \cmidrule{2-3} 
~&\textbf{AVIS} (ours) & \textbf{60.2} \\ 
~& ~~ w/o PALI$^*$    & 47.1 \\
~&~~ w/o Object      & 58.3\\
~&~~ w/o Search  & 55.0\\
\bottomrule
\end{tabular}
\caption{\textbf{Visual Question Answering} results (accuracy) on OK-VQA. The tool PALI$^*$ denotes the frozen multi-task PALI-17B model for both visual question answering and image captioning. Object means object detection, and search means image and text search. }
\label{table:okvqa}
\end{table}

\textbf{Baselines.}
A significant novelty of \framework is the ability to dynamically determine the relevant tools according to different states. To show that this design choice is useful, 
we add a number of baselines that do not contain a LLM-planner for dynamic decision making. 
Instead, they follow a pre-determined sequence to call a list of tools. We propose the following baselines:
\begin{compactitem}
    \item \textbf{baseline-PALM w/ PALI$^*$}, which integrates the captions generated by PALI and the visual answers from PALI VQA. PALI$^*$ denotes the combination of both VQA and captioning tool.
    \item \textbf{baseline-PALM w/ (PALI$^*$ + Object)}, which in addition calls the object detection tool, and then integrates all object data, including products and text detected by OCR.
    \item \textbf{baseline-PALM w/ (PALI$^*$ + Object + Search)}, a model which first selects a relevant object with the help of PALM, then sequentially executes the image search and Google search with the object name. It then calls PALM again to answer the question.
\end{compactitem}
For each of the three baselines, we prepare a few-shot Chain-Of-Thought (COT) prompting~\cite{DBLP:conf/nips/Wei0SBIXCLZ22}, in which the COT prompt guides the model to explain why predictions are made based on the provided information.
Note that these baselines utilize a set of tools in a fixed order, without the capacity for dynamic decision making.

We also evaluate the usefulness of each tool group (i.e., PALI$^*$, Object, and Search) through an ablation study. 
This involves removing each tool group from our framework individually, and assessing the impact on performance. 

\begin{figure}
    \centering
    \includegraphics[width=1\columnwidth]{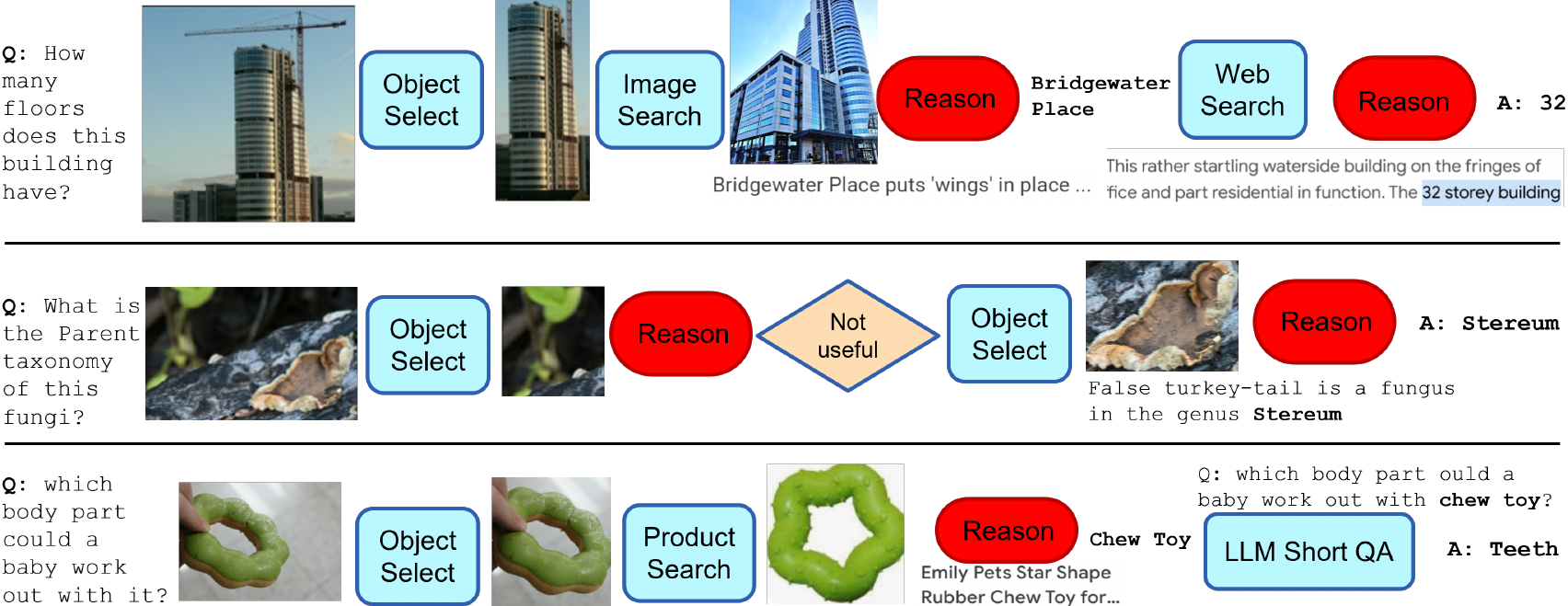}
    \vspace*{-.2in}
    \caption{Examples of \framework's dynamic planning and reasoning procedure for solving visual questions. 
    \vspace*{-.3in}
    }
    \label{fig:example}
\end{figure}

\paragraph{Experimental Results.}
Table~\ref{table:infoseek} presents the results of \framework and other baselines on the Infoseek$_{\text{wikidata}}$ dataset. Infoseek$_{\text{wikidata}}$ is a challenging dataset that requires identifying highly specific entities. 
Even robust visual-language models, such as OFA~\cite{DBLP:journals/corr/abs-2206-08916} and PALI~\cite{DBLP:journals/corr/abs-2209-06794}, fail to yield high accuracy when fine-tuned on this dataset. However, our \framework, without fine-tuning and by leveraging a complete set of tools guided by 10 in-context examples, achieves the accuracy of 50.7 and 56.4 on the unseen entity and question splits, respectively.
This significantly outperforms the fine-tuned results of PALI-17B, which are 16.0 and 20.7, as well as the PALM model augmented with CLIP knowledge, which are 21.9 and 18.6, respectively.

Table~\ref{table:infoseek} also illustrates that our improvements are not solely due to the additional information provided by the external tools, but due to our dynamic decision-making pipeline.
We compare the results of \framework with the three baselines that conduct sequential execution. 
While these baselines do improve the performance, our AVIS framework outperforms the best baseline model by up to 17.3 accuracy. 
Note that AVIS and the baselines use exactly the same set of tools. 
This considerable performance gap clearly shows the clear advantage of our dynamic decision-making design. Furthermore, we show the importance of each tool in the last block of Table~\ref{table:infoseek}.
Removal of any of the tools degrades the overall accuracy.
Among the three tool groups, Object and Search are more important than PALI, as they provide more fine-grained information crucial for the Infoseek dataset. 

We report the OK-VQA experiments in Table~\ref{table:okvqa}. 
\framework with few-shot in-context examples achieves an accuracy of 60.2, higher than most of the existing methods tailored for this dataset, including KAT~\cite{DBLP:conf/naacl/GuiWH0BG22}, ReVIVE~\cite{DBLP:journals/corr/abs-2206-01201} and REVEAL~\cite{DBLP:journals/corr/abs-2212-05221} . 
\framework achieves lower but comparable performance compared to PALI model fine-tuned on OK-VQA. 
This difference, compared to Infoseek, may be attributed to the fact that most QA examples in OK-VQA rely more on commonsense knowledge than on fine-grained knowledge. 
Therefore, it is feasible to encode such generic knowledge in the model parameters and requires less external knowledge. Note that PALI zero-shot VQA model itself achieves 41.6 accuracy, which is significantly higher than in Infoseek, which supports this hypothesis. 
Table~\ref{table:okvqa} also shows that the object detection is less crucial as a tool on this data set, compared to PALI captioning and VQA.
Nonetheless, \framework equipped with all tools achieves the best performance.


\paragraph{Case studies for dynamic decision making.} 
One of the key features of \framework is its ability to dynamically make decisions instead of executing a fixed sequence. Figure~\ref{fig:example} presents three examples of \framework's dynamic planning and reasoning process. They demonstrate the flexibility of AVIS to use different tools at various stages. It is also worth noting that our reasoner design enables \framework to identify irrelevant information, backtrack to a previous state, and repeat the search. For instance, in the second example concerning the taxonomy of fungi, \framework initially makes an incorrect decision by selecting a leaf object. However, the reasoner identifies that this is not relevant to the question, prompting \framework to plan again. This time, it successfully selects the object related to false turkey-tail fungi, leading to the correct answer, Stereum. Some detailed error analysis is shown in Appendix~\ref{sec:error}.

\section{Conclusion}
In this paper, we propose a novel approach that equips the Large Language Models (LLM) with the tree-search to use a variety of tools for answering knowledge-intensive visual questions. Our methodology, anchored in human decision-making data collected from a user study, employs a structured framework that uses an LLM-powered planner to dynamically decide on tool selection and query formation. An LLM-powered reasoner is tasked with processing and extracting key information from the output of the selected tool. Our method iteratively employs the planner and reasoner to leverage different tools until all necessary information required to answer the visual question is amassed.

\textbf{Limitation Statement: } Currently \framework is specifically designed for visual question answering. We aim to extend our LLM-powered dynamic decision-making framework to address other reasoning tasks. Additionally, our current framework depends on a computationally intensive LLM, namely, the PALM model. We are interested in investigating whether this decision-making framework can also be performed by lighter weight language models.



{\small
\bibliographystyle{abbrv}
\bibliography{ref}
}

\newpage
\appendix\definecolor{codegreen}{rgb}{0,0.6,0}
\definecolor{codegray}{rgb}{0.5,0.5,0.5}
\definecolor{codepurple}{rgb}{0.58,0,0.82}
\definecolor{backcolour}{rgb}{0.95,0.95,0.92}

\lstdefinestyle{mystyle}{
    backgroundcolor=\color{backcolour},   
    commentstyle=\color{codegreen},
    keywordstyle=\color{magenta},
    numberstyle=\tiny\color{codegray},
    stringstyle=\color{codepurple},
    basicstyle=\ttfamily\tiny,
    breakatwhitespace=true,     
    alsoletter=\\\{\}\*\[\]\-,
    breaklines=true,                 
    captionpos=b,                    
    keepspaces=true,                 
    numbers=left,                    
    numbersep=1pt,                  
    showspaces=false,                
    showstringspaces=false,
    showtabs=false,                  
    tabsize=2,
}

\lstset{style=mystyle}

\section{Implementation of AVIS workflow}
We implemented AVIS using the code snippet referenced in Code~\ref{code:workflow}. Throughout our experiments, we employed the APIs of Google Search, LENS, PALI, and PALM directly, without the need for additional GPU/TPU computational resources. Tools that didn't require input queries, such as object detection, captioning, and image search, had their results pre-calculated over the two datasets to reduce the time cost. Other services like VQA, text search, and LLM QA were called during runtime.


\section{Comparison to pure Autonomous baseline without Transition Graph}
One of the significant contributions of this paper lies in the use of a transition graph, synthesized from an authentic user study. To underscore the importance of this graph, along with user prompts in facilitating the efficacy of AVIS, we devised a baseline that operates independently of the transition graph. In this scenario, the model, at each timestep, is presented with a comprehensive list of all tools, each paired with a task description. This baseline shares similarities with the recently launched AutoGPT~\footnote{\url{https://github.com/Significant-Gravitas/Auto-GPT}}, BabyAGI\footnote{\url{https://github.com/yoheinakajima/babyagi}} projects, which attempted to utilize LLMs as autonomous agents to select all possible actions available in the web.

The results are show in Table~\ref{tab:graph} on Infoseek WIkiData unseen entity set and OKVQA. Note that this baseline doesn't achieve the number as high as AVIS with the transition graph and user prompts. The key reason for this discrepancy is the global characteristics inherent in the tool list we have. For instance, we typically first address the visual sub-question through object detection and image search, followed by resolving the knowledge component via Google Search and LLM. However, solely relying on the task description, devoid of human behavior as guidance, can result in the model generating unrealistic tools. We will discuss this intuition more in the following sections.

\begin{table}[h]
    \centering
    \begin{tabular}{ccc} \toprule
    Model & Infoseek & OKVQA \\ \midrule
        AVIS w.o/ 
Transition Graph &  38.2 & 47.3 \\
        AVIS w/ Transition Graph & 50.7 & 60.2  \\ \bottomrule
    \end{tabular} 
    \caption{Ablation of AVIS with or without the guidance of Transition Graph} 
    \label{tab:graph}
\end{table}

\begin{figure}[t!]
    \centering
    \includegraphics[width=0.9\columnwidth]{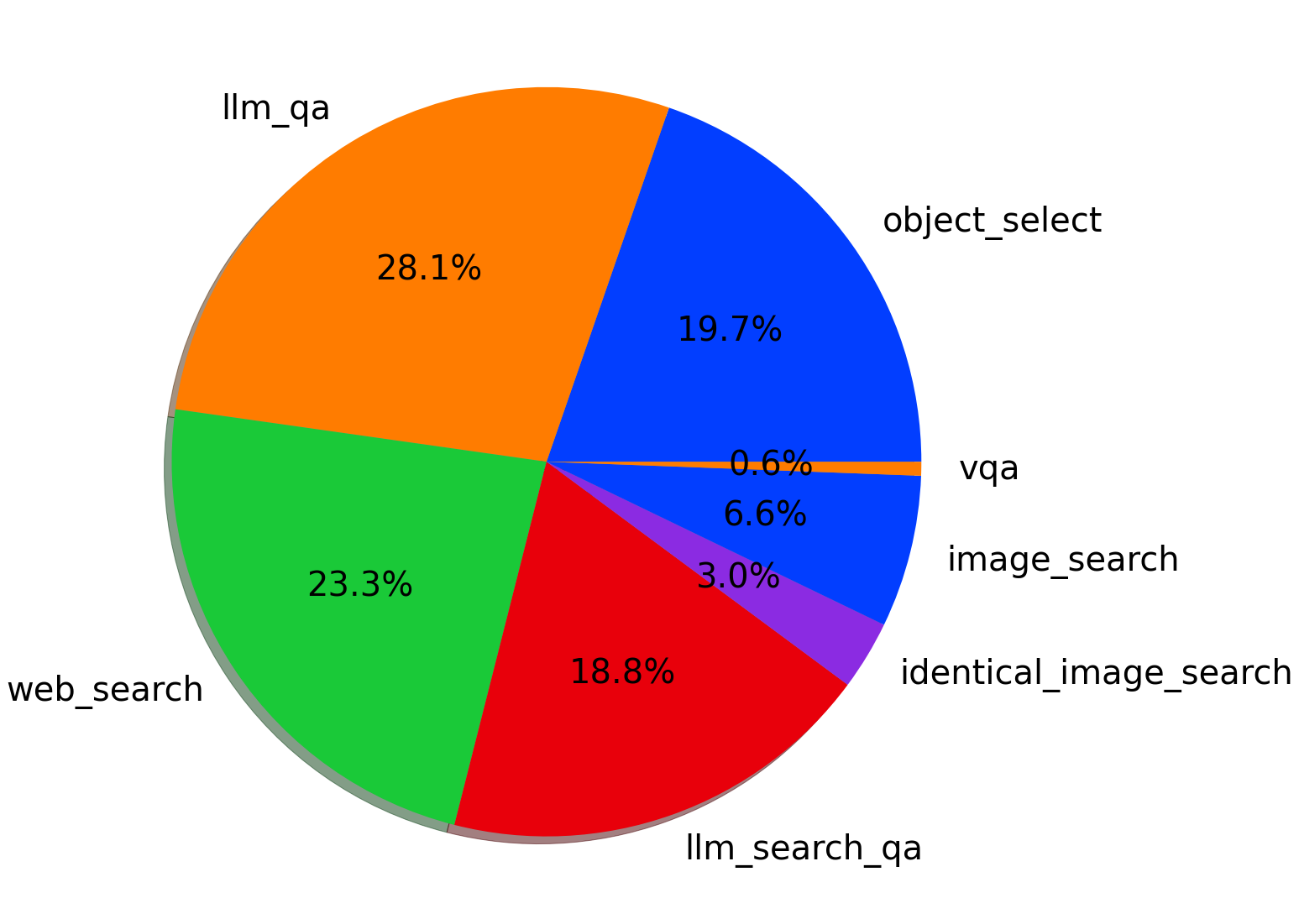}
    \caption{Overall frequency of tool usage on Infoseek dataset.}
    \label{fig:freq}
\end{figure}

\section{Analysis of AVIS's generated tool execution sequence}
We have also conducted an analysis to determine whether common patterns exist within the generated programs of AVIS's predictions.

\begin{figure}[t]
\begin{floatrow}
\ffigbox{
    \includegraphics[width=1.1\columnwidth]{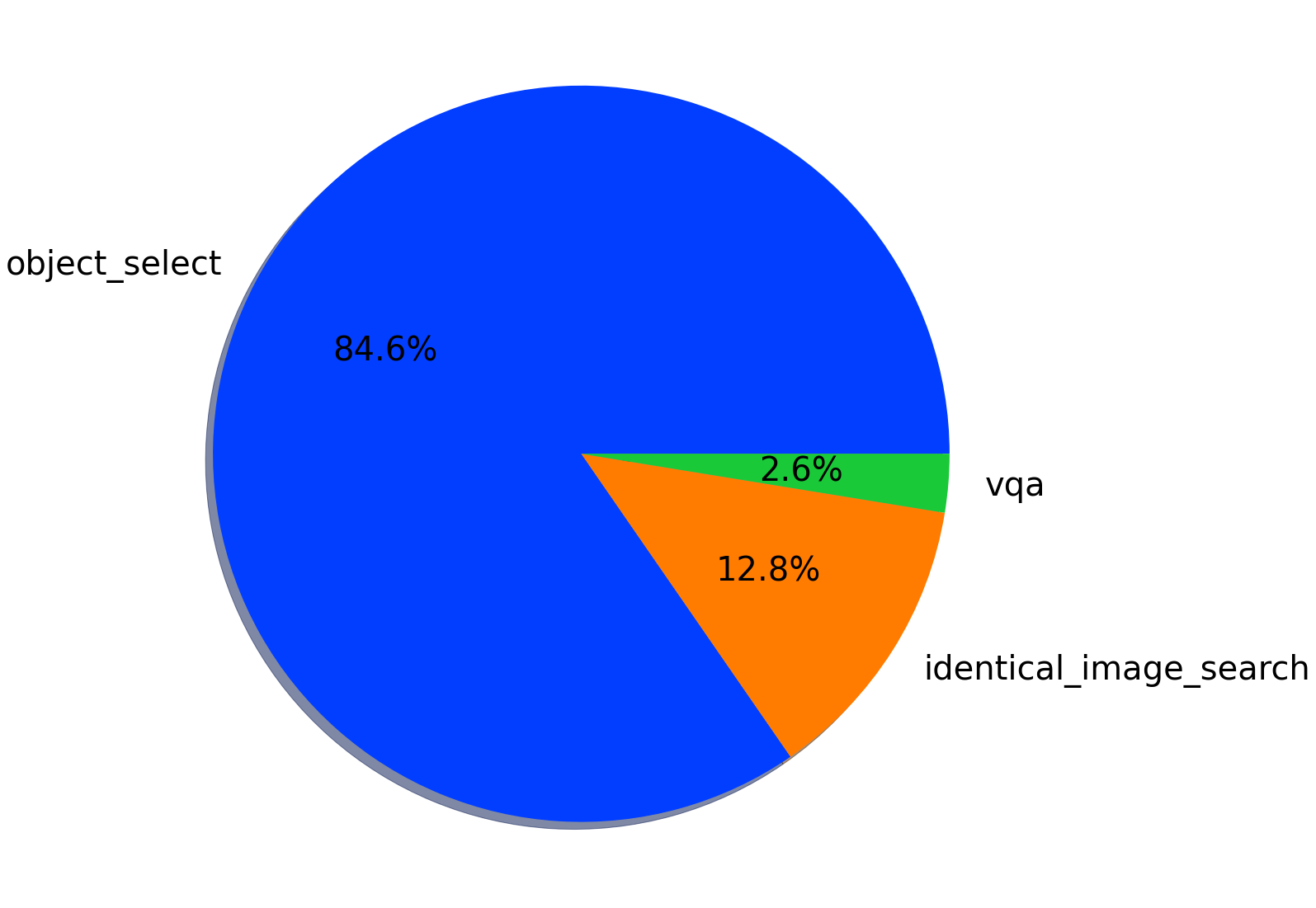}
}{%
   \caption{Frequency of the first used tool.}\label{fig:freq_first}
}
\ffigbox{
    \includegraphics[width=0.8\columnwidth]{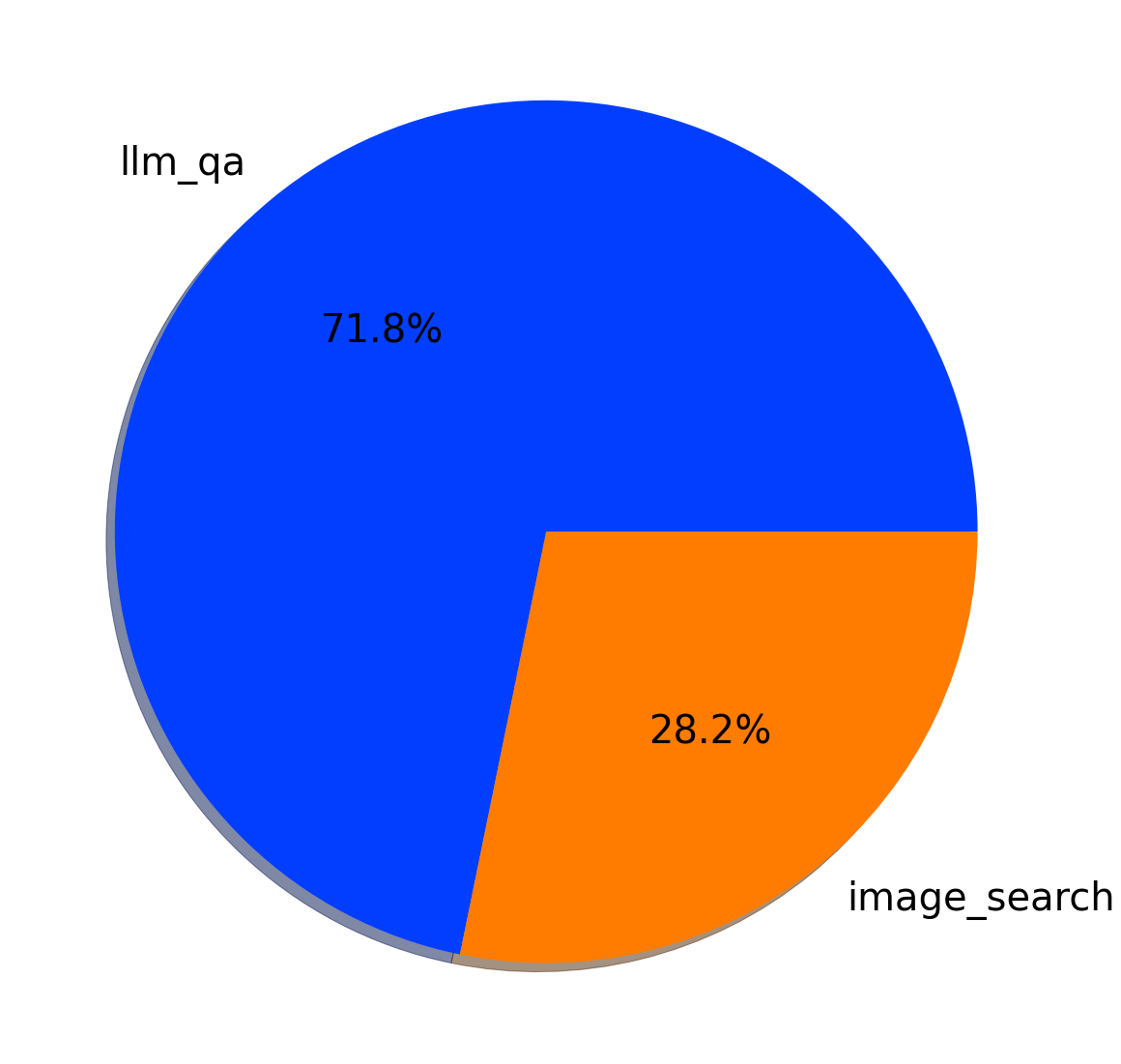}
}{%
   \caption{Frequency of the second used tool.}\label{fig:freq_second}
}
\end{floatrow}
\end{figure}

\begin{figure}[t]
\begin{floatrow}
\ffigbox{
    \includegraphics[width=0.9\columnwidth]{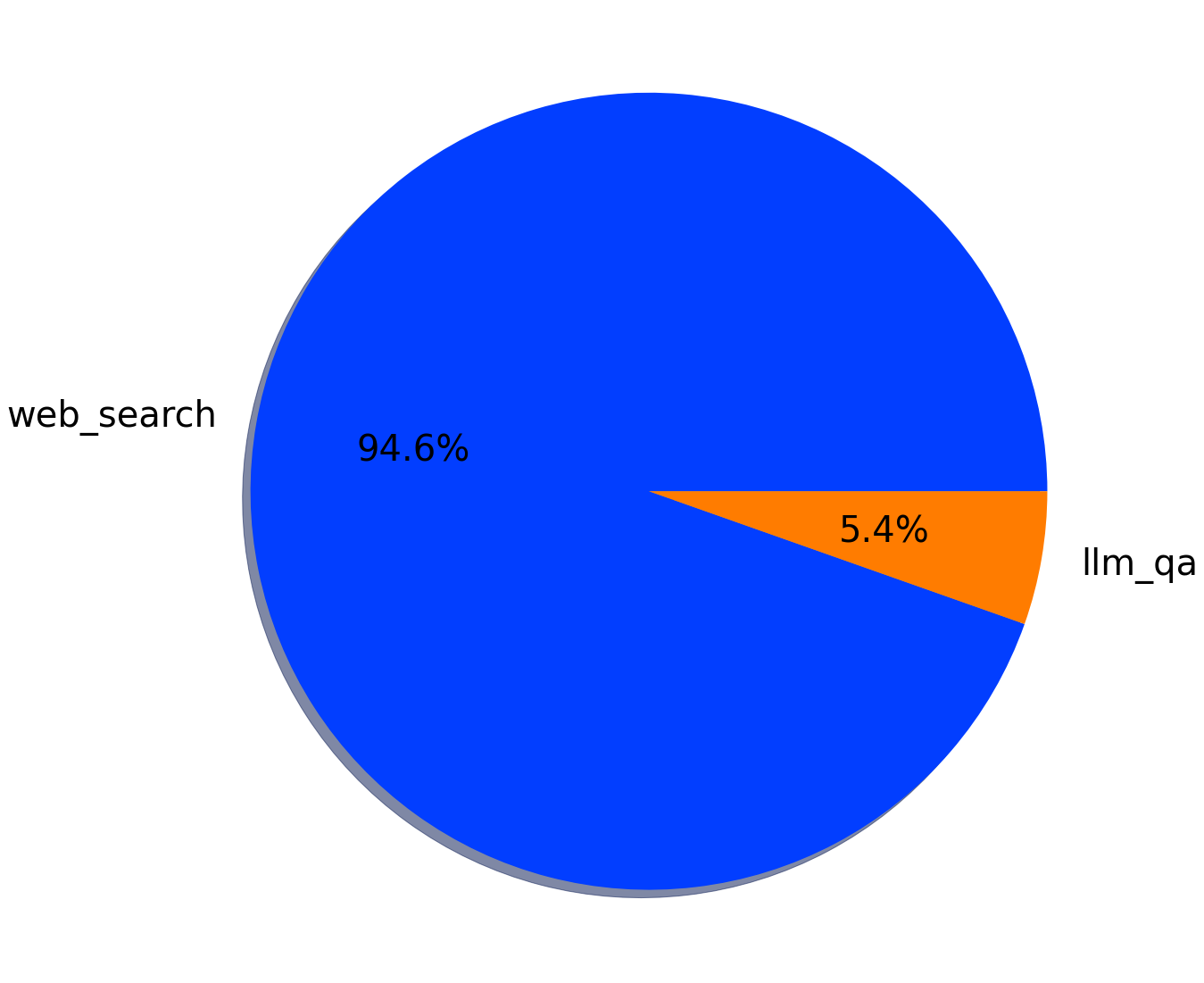}
}{%
   \caption{Frequency of the third used tool.}\label{fig:freq_third}
}
\ffigbox{
    \includegraphics[width=1.0\columnwidth]{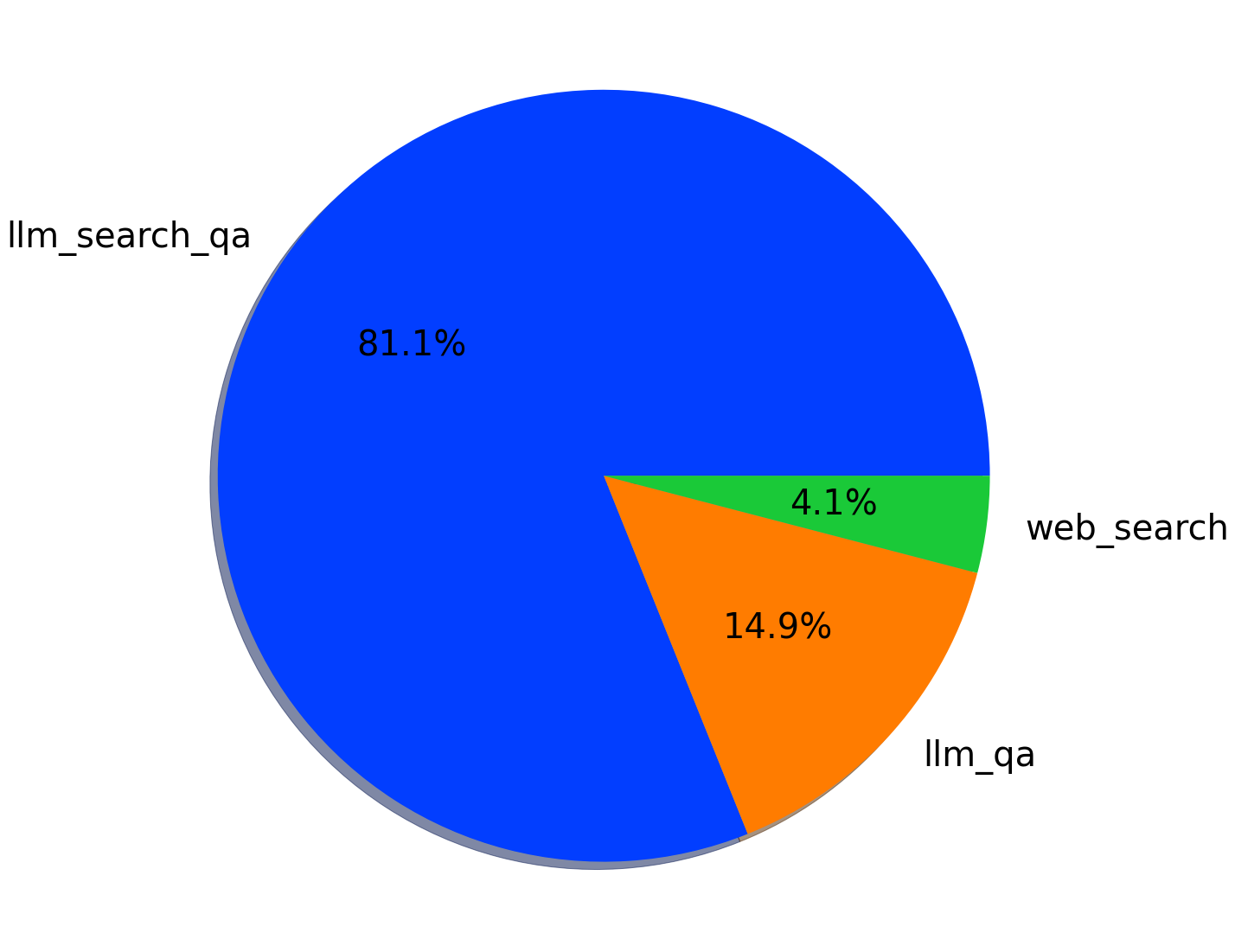}
}{%
   \caption{Frequency of the forth used tool.}\label{fig:freq_forth}
}
\end{floatrow}
\end{figure}

We gathered the tool execution traces for all samples within the Infoseek unseen entity dataset. Initially, we display the frequency of each tool being invoked in Figure~\ref{fig:freq}, followed by a more detailed analysis of the first to fourth most commonly called tools in Figures~\ref{fig:freq_first}-\ref{fig:freq_forth}. As illustrated, the AVIS model, guided by the transition graph and prompts, does not utilize all possible combination of tools, but favors some certain combinations. For instance, as depicted in Fig~\ref{fig:freq_first}, "object select" is utilized more frequently than other tools at the outset. Similarly, as demonstrated in Fig~\ref{fig:freq_third}, during the third step, when the model accumulates the visual answer, it is likely to invoke "web search" to gather additional information.

\begin{figure}[t]
    \centering
    \includegraphics[width=1.0\columnwidth]{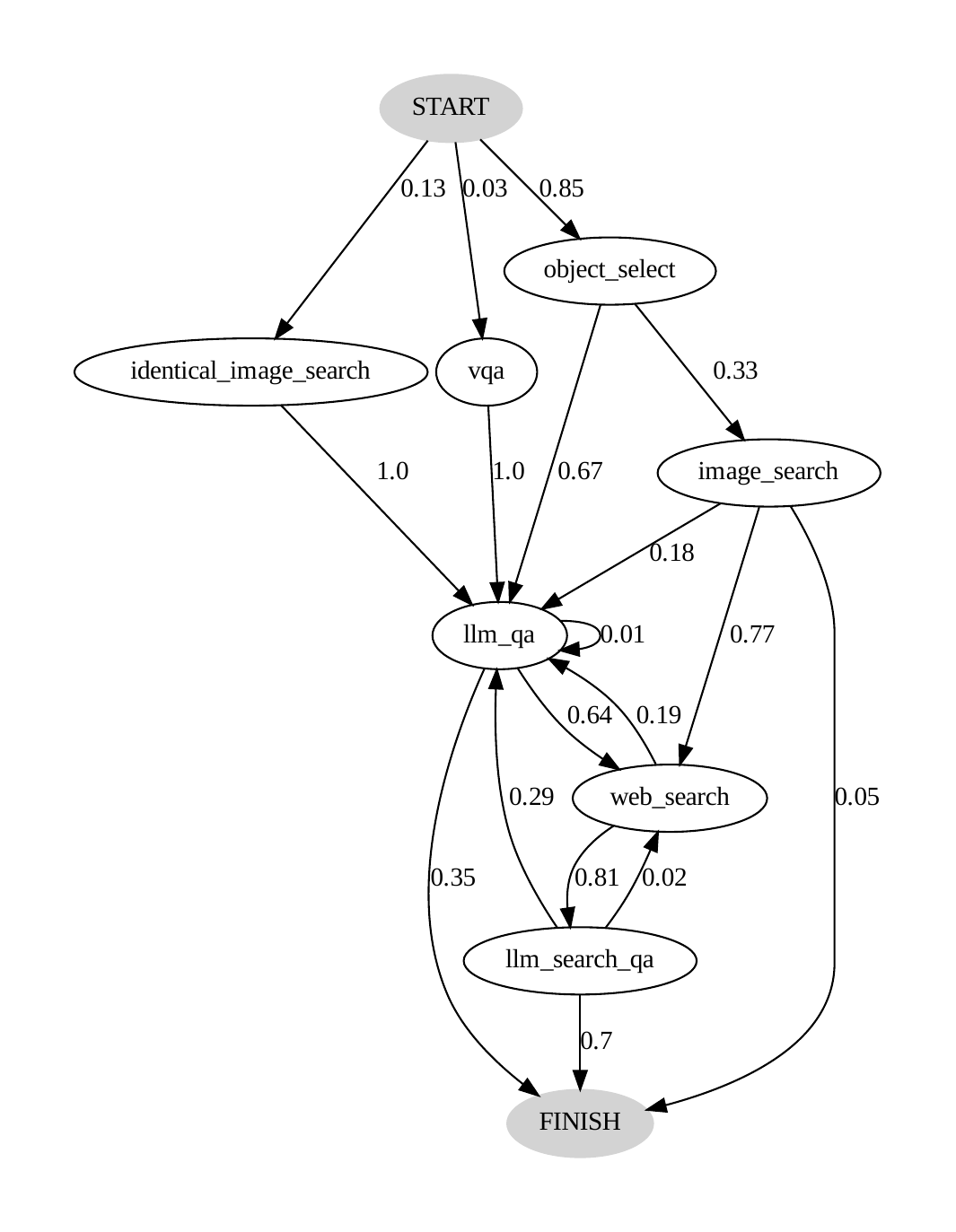}
    \caption{Induced transition frequency graph of AVIS over Infoseek dataset.}
    \label{fig:infoseek_graph}
\end{figure}

\begin{figure}[t]
\begin{floatrow}
\ffigbox{
    \includegraphics[width=1.0\columnwidth]{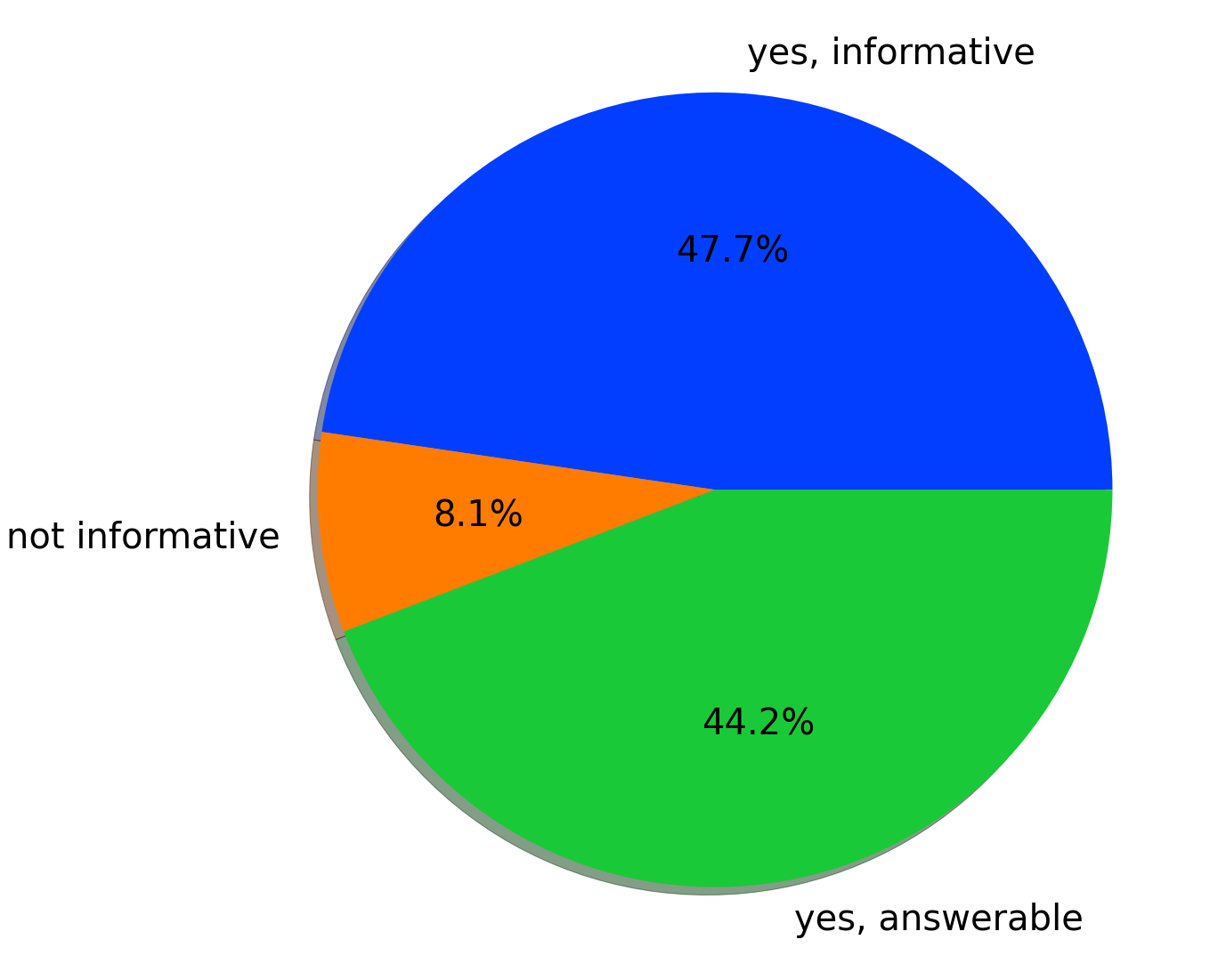}
}{%
   \caption{Overall frequency of judgement by reasoner of AVIS.}
    \label{fig:reason}
}
\ffigbox{
    \includegraphics[width=1.1\columnwidth]{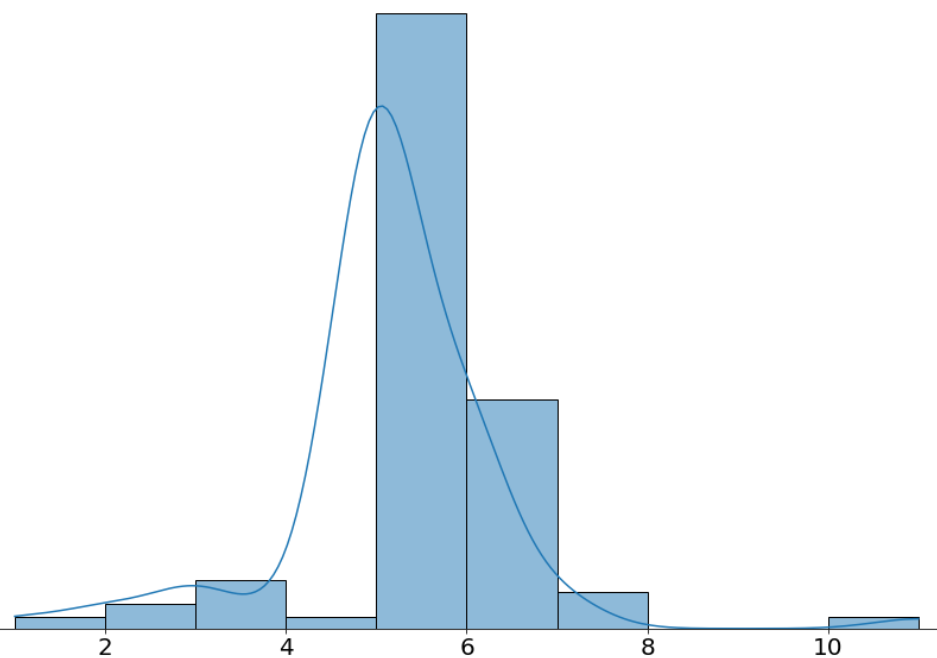}
}{%
   \caption{Length distribution of AVIS's generated action sequences.}\label{fig:length}
}
\end{floatrow}
\end{figure}

We have also calculated the transition probability of the induced graph in Fig~\ref{fig:infoseek_graph}. The structure of this graph differs slightly from the guided transition graph because during actual runtime, the model will not predict some of the edges. Overall, it reveals a clear two-step question-solving pattern. Initially, AVIS gathers sufficient visual information through the use of visual tools such as "object detection," "VQA," or "identical image search," and then employs "LLM QA" to obtain the visual answer. Subsequently, it iteratively calls "web search" and "LLM QA" post-search with a prompt, eventually deriving the final answer. We also present the distribution of the lengths of generated sequences in Figure~\ref{fig:length}. As illustrated, the lengths vary considerably, rather than maintaining a fixed value, with a length of 5 being most common for the generated sequences.

Another intriguing aspect worth exploring is our reasoner component. As explained in the paper, the reasoner evaluates whether the output of each tool is "informative," "not informative," or "answerable". We exhibit the overall frequency of these predictions in Figure~\ref{fig:reason}. As shown, the model tends to classify most of the outputs as either informative or answerable. However, approximately 8.1\% of returned entries are deemed "not informative," in which case AVIS would backtrack to select alternative actions. We further demonstrate a few examples of different choices in Table~\ref{tab:example_reasoner}.


\begin{table}
\scriptsize
\begin{tabular}[t]{c|c|c|c} \toprule
Question & Action & Returned Content & Output of Reasoner \\ \midrule
\multicolumn{1}{p{2cm}|}{where is the island located in?}&\multicolumn{1}{p{1cm}|}{image search} & \multicolumn{1}{p{8cm}|}{Retrievals: [willow - Students | Britannica Kids | Homework Help (score=57.2), Silky Willow Live Stakes For Sale | Buy Live Stakes Wholesale (score=55.3), ...]} & \multicolumn{1}{p{2cm}}{not informative} \\ \midrule
\multicolumn{1}{p{2cm}|}{Who or what maintains this place?}&\multicolumn{1}{p{1cm}|}{LLM QA} & \multicolumn{1}{p{8cm}|}{the history of trafalgar square. the most relevant knowledge entry about trafalgar square is "Trafalgar Square is the focal point for protests, revelry and victory celebrations, however it was only in 1835 that the area became known by this name".
From the result, we know the history of trafalgar square is trafalgar square is the focal point for protests, revelry and victory celebrations, however it was only in 1835 that the area became known by this name. Terefore, the predicted answer is trafalgar square is the focal point for protests , revelry and victory celebrations , however it was only in 1835 that the area became known by this name.} & \multicolumn{1}{p{2cm}}{not informative} \\ \midrule
\multicolumn{1}{p{2cm}|}{What is the total quantity of produced items for this type of aircraft?}&\multicolumn{1}{p{1cm}|}{image search} & \multicolumn{1}{p{8cm}|}{Retrievals: [high confidence: Condor bietet neue Langstrecke nach Brasilien an - AERO International (score=97.3), high confidence: Air safety incidents for aircraft registration D-ABOE - AeroInside (score=95.0), ...]} & \multicolumn{1}{p{2cm}}{yes, answerable} \\ \midrule
\multicolumn{1}{p{2cm}|}{what is the name of this mountain?}&\multicolumn{1}{p{1cm}|}{image search} & \multicolumn{1}{p{8cm}|}{Object: [
  a view of a castle in the distance . There are trees and buildings present at the bottom of this image. We can see a hill and a tower in the middle of this image. We can see the sky in the background. (Caption, whole image), Mount of Olives (ridge): The Mount of Olives or Mount Olivet is a mountain ridge east of and adjacent to Jerusalem's Old City (score=88.6), Mount Zion (peak): Mount Zion is a hill in Jerusalem, located just outside the walls of the Old City (score=79.0)]} & \multicolumn{1}{p{2cm}}{yes, informative} \\ \bottomrule
\end{tabular}
\caption{Several examples of API execution results and the reasoner's justification.}
\label{tab:example_reasoner}
\end{table}

\section{Dataset Details}


\textbf{Infoseek}\footnote{\url{https://open-vision-language.github.io/infoseek/}} is a Visual Question Answering (VQA) dataset, specifically geared towards information-seeking questions that cannot be answered merely through common sense knowledge. This dataset was curated by initially gathering human-annotated questions, which were then automatically integrated with existing visual entity recognition datasets and Wikidata to generate complex question-answer pairs. At the time of submission, we only have access to its wikidata split. Here we also report the results on human split in Table~\ref{table:infoseek}.

\textbf{OK-VQA}\footnote{\url{https://okvqa.allenai.org/}} is another VQA dataset, unique in its requirement for the application of external knowledge that transcends the information directly visible in the input images. The creation of this dataset involved crowdsourced workers who were tasked with annotating complex questions, drawing upon the extensive knowledge resources available on Wikipedia.

\begin{table}[t]
\small
\centering
\begin{tabular}{lcc}
\toprule  
\textbf{Model}  & Unseen Entity & Unseen Question \\             
\midrule
PALM (Q-only, few-shot)                   & 6.6     & 4.8                           \\
OFA (fine-tune)                           & 2.9    &   6.2                        \\
PALI (fine-tune)                      & 5.9    &   13.3                      \\ 
PALM w/ CLIP (few-shot + external knowledge)                  & 14.9   &    15.6                       \\ 
FiD w/ CLIP (fine-tune + external knowledge)                  & \underline{17.6}    &    \underline{18.9}                      \\ 
\midrule
\textbf{\framework} (ours, few-shot)    & \textbf{31.4}   &   \textbf{33.6}\\
\bottomrule
\end{tabular}
\caption{\textbf{Visual Question Answering} results (accuracy) on Infoseek$_{\text{human}}$. The first four rows are results from their paper that do not use external knowledge, and the next two are from their paper that use CLIP as knowledge source.  
}\label{table:infoseek}
\end{table}

\section{Prompt Examples}

Below we show different prompt examples to support our AVIS workflow. First is the prompts for planning, which selects which tool to use and what query to send. It is consists of a overall task descriptions and many real examples showing at which circumstances real users select this tool.

\begin{lstlisting}[language=Python,caption={\small{Planner prompt skeleton and Task instructions
}},label={code:task},floatplacement=H]
planner_prompt = 
"""You goal is to answer the following query: %s.

To answer it, you will be provided with the following tools:
%s

Please make the decision based on the current context.

%s
Query: %s
Context: %s
Action: \n
"""

task_instructions = {
'vqa': 
    'You will ask simple question about this image to a external QA module. Please use this when the input query is very straightforward and simple.',\
'object_select': 
    'You will select one of the object we detect to dig further. Please use when the question asks about a specific object.',\
'identical_image_search': 
    'You will see captions of all images identical to the given image. Please use when the question asks about the whole image instead of a part.',\
'image_search': 
    'You will see captions of all images similar to this object. Please use when you need more information.',\
'web_search': 
    'You will send question to Google Search to get knowledge. Please use when the current query requires extra knowledge',\
'llm_qa': 
    'You will send question to a QA module. Please use this when the input query is simple and contain common-sense knowledge'
}
\end{lstlisting}

\newpage
\begin{lstlisting}[language=Python,caption={\small{Planning Prompts Example}},label={code:task},floatplacement=H]
vqa_plan_prompts = [
"""Query: what is the train carrying?
Context: [
  a train traveling down train tracks next to a forest . There are four trains on the railway track. In the background there are trees,poles and sky. (Caption, whole image)
  Extracted Text: BNSF (score=100.0),
  BNSF Railway: BNSF Railway is one of the largest freight railroads in North America (score=89.3),
]
Action: vqa
""",\
"""Query: What is the girl wearing on her legs?
Context: [
  a woman standing in a field putting on a coat . There is a woman standing on the ground. This is grass and there are plants. In the background we can see some trees and this is sky. (Caption, whole image)
]
Action: vqa
""",\
"""Query: what color is the bus?
Context: [
  a double decker bus parked in front of a building . There is a double decker bus on the road and this is snow. Here we can see a pole, light, trees, and houses. In the background there is sky. (Caption, whole image)
  Extracted Text: ENVIRO400 (score=100.0),
  Extracted Text: Les Miserables (score=100.0),
  Query Suggestion: les miserables (score=100.0),
  Volvo Olympian: The Volvo Olympian was a rear-engined 2-axle and 3-axle double decker bus chassis manufactured by Volvo at its Irvine, Scotland factory (score=88.5),
  Alexander Dennis Enviro400: The Alexander Dennis Enviro400 is a twin-axle low-floor double-decker bus that was built by the British bus manufacturer Alexander Dennis between 2005 and 2018 (score=85.4),
]
Action: vqa
""",\
"""Query: what is the person doing?
Context: [
  two people sitting on the floor opening presents . There are sofas on the sofas there are pillows, here there is table, on the table there are plants and other objects, here there are two persons sitting on the ground, gift boxes, dog and this is floor. (Caption, whole image)
]
Action: vqa
"""
]
object_select_plan_prompts = [
"""Query: what is the name of this building?
Context: [
  a group of people that are standing in front of a building . There is a building in the left corner which has few people standing in front of it and there is a fire hydrant in the right corner and there is a street light pole beside it. (Caption, whole image)
  Query Suggestion: Alcatraz Warden's House San Francisco (score=100.0),
  Alcatraz Island (historic_site): Alcatraz Island is a small island 1 (score=91.9),
  Warden's House: The Warden's House was the home of the wardens of the federal penitentiary on Alcatraz Island, off San Francisco (score=78.1),
]
Action: object_select
""",
"""Query: what is the island?
Context: [
  a view of a mountain from a cable car . There is a ropeway. Behind that there are trees and hills. (Caption, whole image)
  Ngong Ping 360 (gondola_lift_station): Ngong Ping 360 is a bicable gondola lift on Lantau Island in Hong Kong (score=91.8),
  Tian Tan Buddha (monument): The Big Buddha is a large bronze statue of Buddha, completed in 1993, and located at Ngong Ping, Lantau Island, in Hong Kong (score=79.0),
]
Action: object_select
""",
"""Query: what is the name of this place?
Context: [
  a cemetery with a building in the background . There is a road and there are many atoms and trees beside it and there is a building in the right corner. (Caption, whole image)
]
Action: object_select
""",
"""Query: what is the name of this bird?
Context: [
  a bird sitting on top of a lush green hillside . There is a bird on the grassland in the foreground area of the image and the background is blurry. (Caption, whole image)
  Atlantic puffin (type_of_bird): The Atlantic puffin, also known as the common puffin, is a species of seabird in the auk family (score=73.2),
  Horned puffin (type_of_bird): The horned puffin is an auk found in the North Pacific Ocean, including the coasts of Alaska, Siberia and British Columbia (score=73.2),
  Puffins (type_of_bird): Puffins are any of three species of small alcids in the bird genus Fratercula (score=73.2),
  Fraterculini (score=48.8),
  Auk (type_of_bird): An auk or alcid is a bird of the family Alcidae in the order Charadriiformes (score=11.8),
]
Action: object_select
"""
]
identical_image_search_plan_prompts = [
"""Query: what is the name of this building?
Context: [
  a group of people that are standing in front of a building . There is a building in the left corner which has few people standing in front of it and there is a fire hydrant in the right corner and there is a street light pole beside it. (Caption, whole image)
  Query Suggestion: Alcatraz Warden's House San Francisco (score=100.0),
  Alcatraz Island (historic_site): Alcatraz Island is a small island 1 (score=91.9),
  Warden's House: The Warden's House was the home of the wardens of the federal penitentiary on Alcatraz Island, off San Francisco (score=78.1),
]
Action: identical_image_search
""",
"""Query: what is the aircraft?
Context: [
  a fighter jet sitting on top of an airport tarmac . There is a plane and missiles on the ground. At the left a person is standing wearing a cap. (Caption, whole image)
  Extracted Text: AIRLINERS.NET (score=100.0),
  Query Suggestion: airliners.net (score=100.0),
  Airliners: Airliners (score=74.8),
  British Aerospace Hawk 200: The British Aerospace Hawk 200 is a single-seat, single engine light multirole fighter designed for air defence, air denial, anti-shipping, interdiction, close air support, and ground attack (score=74.8),
  product: Airfix BAE Hawk T1 1:72 (score=0.0),
  product: Rolls-royce Adour In The Hawk / Bae Hawk 200 . Pdf/download (score=0.0),
]
Action: identical_image_search
""",
"""Query: what is the name of this place?
Context: [
  a row of pillars sitting next to a dirt road . There is a building and this is plant. Here we can see pillars and a sky. (Caption, whole image)
  Query Suggestion: Palmyra Archaeology (score=100.0),
  Great Colonnade at Palmyra (ancient_roman_architecture_structure): The Great Colonnade at Palmyra was the main colonnaded avenue in the ancient city of Palmyra in the Syrian Desert (score=90.3),
]
Action: identical_image_search
""",
"""Query: what is the name of this lake?
Context: [
  a view of a river surrounded by mountains . There are trees in the right corner and there is a river and mountains in front of it. (Caption, whole image)
  Monte Bre (peak): Monte Bre is a small mountain east of Lugano on the flank of Monte Boglia with a view of the bay of Lugano and the Pennine Alps and the Bernese Alps (score=85.5),
  product: Top Searched (score=0.0),
]
Action: identical_image_search
"""
]
action_prompt_dict = {'vqa': vqa_plan_prompts, 'object_select': object_select_plan_prompts, 'identical_image_search': identical_image_search_plan_prompts, 'image_search': image_search_plan_prompts, 'web_search': web_search_plan_prompts, 
'llm_qa': llm_qa_plan_prompts}
\end{lstlisting}

We then show how AVIS decompose question into a visual sub-question and a knowledge sub-question. This is done at beginning to guide later tool usage.
\begin{lstlisting}[language=Python,caption={\small{Question Decomposition Prompts
}},label={code:task},floatplacement=H]
question_decomposition_prompt = """
    Read the following question for a given image. Decompose the question into two sub-questions. 

    The first will ask information about the image, and the second requires reasoning over the textual knowledge.
    In the second question, we use # to denote the answer of the first question.


    Question: what chemical makes the vegetable orange?
    Visual: which orange vegetable is shown?
    Knowledge: chemical makes # orange?
    
    
    Question: How long can their horns grow?
    Visual: which animals are shown?
    Knowledge: How long can #'s horns grow?
    
    
    Question: What is a competition for these animals called?
    Visual: which animals are shown?
    Knowledge: competition for #?
    
    
    Question: What is the name of the ancient greek sport that evolved into the sport featured above?
    Visual: which sport is played?
    Knowledge: name of the ancient greek sport that evolved into #?
    
    
    Question: Which food item here has the most protein?
    Visual: what are the food items shown?
    Knowledge: Which food item of # has the most protein?
    
    
    Question: How many calories are in this meal?
    Visual: what are the food items shown?
    Knowledge: calories in #?
    
    
    Question: What type of sandwich is this?
    Visual: which type of sandwich is shown?
    Knowledge: #
    
    Question: What is the name of the restaurant where this was served?
    Visual: which food items are served?
    Knowledge: restaurant where # was served?
    
    
    Question: What genus of bird is flying here?
    Visual: what genus of bird is flying?
    Knowledge: #
    
    
    Question: What is the main ingredient in this food?
    Visual: which food is shown?
    Knowledge: main ingredient in #?
"""
\end{lstlisting}

Below are several examples to help AVIS learns how to select the most suitable object ID.
\begin{lstlisting}[language=Python,caption={\small{Object Select Prompts
}},label={code:task},floatplacement=H]
object_select_prompt = """
    Please think step by step. In the following, you will be given a "Query", a list of "Objects". 

    Your task is to predict the object #ID that is mostly relevant to answer the querys. Please generate the detailed explanation why you select this object, and then output ID in "Object #ID". 


Query: which city is this place?
Object #0 [
  a row of pillars sitting next to a dirt road . There is a building and this is plant. Here we can see pillars and a sky. (Caption, whole image)
  Query Suggestion: Palmyra Archaeology (score=100.0),
  Great Colonnade at Palmyra (ancient_roman_architecture_structure): The Great Colonnade at Palmyra was the main colonnaded avenue in the ancient city of Palmyra in the Syrian Desert (score=90.3),
]
Object #1 [
  a green plant sitting next to a brick wall . There is a plant and this is wall. And there is a sky. (Caption, center)
  Date palm (type_of_palm_trees): Phoenix dactylifera, commonly known as date palm, is a flowering plant species in the palm family, Arecaceae, cultivated for its edible sweet fruit called dates (score=81.7),
]
Object #2 [
  a wicker basket sitting on top of a rock . There is a blur image of a rock. (Caption, lower right)
]
Output: The query asks about the city of the place. Only Object #0 (whole image) mentions city name Palmyra, which is an acient city. Also, Object #0 contains Query Suggestion "Palmyra Archaeology".
Therefore, the predicted Object #ID is 0.


Query: where is this place?
Object #0 [
  a view of a valley surrounded by mountains . There are hills and this is grass. Here we can see trees and this is sky. (Caption, whole image)
]
Object #1 [
  a view of a lush green hillside with trees . There is a house on the rock and there are few plants beside it and there is a greenery ground in the background. (Caption, center)
  Monterey Pine (type_of_conifers): Pinus radiata, the Monterey pine, insignis pine or radiata pine, is a species of pine native to the Central Coast of California and Mexico (score=49.1),
  European rabbit (type_of_leporids): The European rabbit or coney is a species of rabbit native to the Iberian Peninsula, western France, and the northern Atlas Mountains in northwest Africa (score=31.3),
]
Object #2 [
  a green plant growing on a rocky surface . There is a blur image of trees and rocks. (Caption, lower center)
  product: GreenView Fairway Formula Seed Success Paillis biodegradable avec engrais Sac de 4,5 kg Couvre 200 m2 (score=0.0),
]
Object #3 [
  a rocky hillside with lots of green vegetation . There are trees and this is rock. (Caption, lower left)
  Willow: Willows, also called sallows and osiers, of the genus Salix, comprise around 350 species of typically deciduous trees and shrubs, found primarily on moist soils in cold and temperate regions (score=31.3),
  Tamarisk: The genus Tamarix is composed of about 50-60 species of flowering plants in the family Tamaricaceae, native to drier areas of Eurasia and Africa (score=26.8),
]
Output: The query asks about the location of this place. Although these entries doesn't explicitly contain location name, but Object #1 (center) contains Monterey Pine and European rabbit, which might hint the location later.
Therefore, the predicted Object #ID is 1.
"""
\end{lstlisting}

Below are the prompts to extract answer from objects and extracted captions of similar images.
\begin{lstlisting}[language=Python,caption={\small{Reason Prompt (Visual Question)
}},label={code:task},floatplacement=H]
reason_vqa_prompt = """
Please think step by step. In the following, you will be given:

- Query: The query to be asked.
- Think: Why the following knowledge is retrieved.
- Entity: A list of entities that describe the object.
- Retrievals: A list of web documents that are similar to the object. If there's "high confidence", it's very important.

Your task is to predict a short answer to the query based on the provided information. You need to first identify which knowledge entry is mostly relevant, and then extract the answer from the knowledge. 
Rely on Object information more, and if there contains "Query Suggestion", try to use it. Otherwise, if a information appears lots of time, there's a higher chance it's the answer. 
After explaining your decision choice, saying "Answer is" and appending your predicted short answer. Please also generate the type of the answer after a comma.
If you are uncertain about the answer, especially when the knowledge is irrelevant to the query, say "cannot be answered". Do not generate the answer not inside the provided knowledge.



Query: what is this building?
Think: object  (whole image) contains stockholm city hall, which is the seat of stockholm municipality in stockholm, sweden.
Object: [
  Stockholm City Hall (city_hall): Stockholm City Hall is the seat of Stockholm Municipality in Stockholm, Sweden (score=96.1),
  Bla Hallen (banquet_hall): The Blue Hall is the main hall of the Stockholm City Hall best known as the banquet hall for the annual Nobel Banquet, and also used for state visits, student balls, jubilees and other large events (score=79.0),
]
Retrievals: [
  high confidence: City Hall - Blue Hall (1) | Stockholm (2) | Pictures | Sweden in Global-Geography (score=47.8),
  high confidence: le salon bleu a city hall (salle de remise des prix nobel) - Picture of Stockholm, Stockholm County - Tripadvisor (score=47.7),
]

Output: The query asks about the building. From both Object and Retrievals, there are mentions about Stockholm City Hall and Blue Hall. As Stockholm City Hall contains Blue Hall, the answer shall be Stockholm City Hall.
Therefore, the predicted answer is Stockholm City Hall.


Query: which sport is played?
Think: Object shows a snail sitting on top of a tennis ball.
Object: [
  Cantareus apertus (type_of_gastropods): Cantareus apertus, commonly known as the green garden snail, is a species of air-breathing land snail, a terrestrial pulmonate gastropod mollusc in the family Helicidae, the typical snails,
  Garden snail (type_of_gastropods): Cornu aspersum, known by the common name garden snail, is a species of land snail in the family Helicidae, which includes some of the most familiar land snails,
  Helix aspersa aspersa (type_of_gastropods),
  Slug: Slug, or land slug, is a common name for any apparently shell-less terrestrial gastropod mollusc,
  Snail: A snail is a shelled gastropod,
]
Retrievals: [
  2019 NEWBIE Competition Winner Steven Ryan, Snail Farming - YouTube,
  Alive specimens. a. Megalobulimus ovatus (CMIOC 11136), b. Thaumastus... | Download Scientific Diagram,
  Brown garden snail > Manaaki Whenua,
  Common garden snail and baby,
  Easy Everyday Food for Garden Snails - Ask the plantician,
  Green Life Soil: Natural pest & disease control in a winter garden,
  Helminthoglyptinae - Wikipedia,
  Hydrosalpingitis in broilers - Veterinaria Digital,
  Master Gardener: Protecting squash and cucumbers from slugs and snails - Press Enterprise,
  Mother Baby Blue Snails On Phalaenopsis Stock Photo 530400856 | Shutterstock,
]

Output: The query asks about sport. From both entities and retrievals, they only talk about snail, and there is no information about which sport is played.
Therefore, given the provided information, this query cannot be answered.


Query: which sport is played?
Think: object , object , and object all contain people playing basketball. however, object is the only one that contains a group of women playing basketball.
therefore, the predicted object #id is 0.
Retrievals: [
  08.07.2011 Zanele Mdodana of South Africa in action during the Quarter-finals between New Zealand and South Africa, Mission Foods World Netball Championships 2011 from the Singapore Indoor Stadium in Singapore Stock Photo - Alamy,
  55 Brazilian Handball Team Images, Stock Photos & Vectors | Shutterstock,
  :::Malawi High Commission:::,
  Amanda Mynhardt Photostream | Netball, Netball singapore, Netball south africa,
  Australia pass Malawi test with flying colours at Netball World Cup | Netball World Cup 2019 | The Guardian,
  Australia's Jo Weston (second left) and Barbados' Latonia Blackman in action during the Netball World Cup match at the M&S Bank Arena, Liverpool Stock Photo - Alamy,
  Birmingham 29795 World Netball Championships Final Editorial Stock Photo - Stock Image | Shutterstock,
  Bridget kumwenda malawi netball hi-res stock photography and images - Alamy,
  England V Australia International Netball Series Photos and Premium High Res Pictures | Netball, Netball quotes, Inspirational women,
  File:Xx0992 - Madrid basketball Donna Burns - 3b - Scan.jpg - Wikimedia Commons,
]

Output: The query asks about which sport is played. From retrievals, there exist many mentions about netball, and mentions that they are played by women.
therefore, the predicted answer is women netball.



Query: what is the name of the insect?
Think: only object  (while image) mentions the name of the insect, western tiger swallowtail.
Object: [
  Query Suggestion: Western Tiger Swallowtail (score=100.0),
  Canadian tiger swallowtail (type_of_lepidoptera): Papilio canadensis, the Canadian tiger swallowtail, is a species of butterfly in the family Papilionidae (score=78.4),
  Eastern tiger swallowtail (us_state_butterfly): Papilio glaucus, the eastern tiger swallowtail, is a species of butterfly native to eastern North America (score=78.4),
]
Retrievals: [
  high confidence: kupu-kupu - Wiktionary (score=100.0),
  high confidence: Top Spots for Nature Watching and Birding | VisitMaryland.org (score=100.0),
  high confidence: File:Eastern Tiger Swallowtail Papilio glaucus on Milkweed 2800px.jpg - Wikimedia Commons (score=99.8),
  high confidence: Photographing Butterflies - Life in the Finger Lakes (score=97.8),
]

Output: The query asks about the name of the insect. From Object, it contains a very informative "Query Suggestion: Western Tiger Swallowtail".
Therefore, the predicted answer is Western Tiger Swallowtail.

"""
\end{lstlisting}

Below are prompts AVIS extract answer from search results:
\begin{lstlisting}[language=Python,caption={\small{Reason Prompt (Knowledge Question)}},label={code:task},floatplacement=H]
reason_qa_prompt = """
    Please think step by step. In the following, you will be given a "Query", and a list of "Knowledge" from Google Search related to this query. 

    Your task is to predict a short answer to the query based on the provided information. You need to first identify the most relevant knowledge entry, and then predict a short answer based on the knowledge. If a information appears lots of time, there's a higher chance it's the answer. 

    After explaining your decision choice, saying "Answer is" and appending your predicted answer. 
    If you are uncertain about the answer, especially when the knowledge is irrelevant to the query, say "cannot be answered". Do not generate the answer not inside the provided knowledge.


Query: What chemical makes carrot orange? 
Knowledge: [
Title: How did carrots become orange? - The Economist
Content: High Confidence Response: carotenoids.

Context: The chemical compounds that give carrots their vivid colour, carotenoids, are usually used by plants that grow above ground to assist in the process of photosynthesis.

Title: 
Content: carotenoids

The chemical compounds that give carrots their vivid colour, carotenoids, are usually used by plants that grow above ground to assist in the process of photosynthesis.

Title: Can Eating Too Many Carrots Make Your Skin Turn Orange? | Britannica - Encyclopedia Britannica
Content: Maybe not! Carrots and other orange fruits and vegetables are rich in a pigment known as beta-carotene. In humans, this pigment is converted to vitamin A by specialized cells in the small intestine. When high levels of beta-carotene are consumed, not all of the pigment is converted to vitamin A.
Fortunately, the skin discoloration fades when the diet is changed and the levels of beta-carotene in the blood decline.

Title: Why are carrots orange? | Ask Dr. Universe | Washington State University
Content: Orange carrots are packed with chemicals called carotenoids-specifically, beta-carotene. Your body turns beta-carotene into vitamin A, which helps you grow and protects you from getting sick. Beta-carotene isn't just nutritious. It's also loaded with orange pigment.
That's why vegetables with lots of beta-carotene-like sweet potatoes, squash, and pumpkins-share the same color. But what about that rainbow of other carrot colors? They have their own special qualities, too. Purple carrots get their color from 
]
Output: The query asks about chemical that makes carrot orange. Because there's one high confidence result, the most relevant knowledge entries about such chemical is "High Confidence Response: carotenoids."
From this result we know the chemical shall be carotene.
Therefore, the predicted answer is carotene.



Query: What is the name of the drainage basin of ounasjoki?
Knowledge: [
Title: Ounasjoki - Wikipedia
Content: It is also the largest river entirely within its borders. Ounasjoki is approximately 299.6 kilometres (186.2 mi) in length, and the catchment area is 13,968 square kilometres (5,393 sq mi), 27% of the Kemijoki catchment area.
Tributaries

- Nakkalajoki.
- Kakkalojoki.
- Syva Tepastojoki.
- Loukinen.
- Meltausjoki.
Course. The Ounasjoki originates at Ounasjarvi lake in Enontekio. It flows first eastwards through Perilajarvi lake and turns south after some seven kilometres. The river then follows southern-sou

Title: DRAINAGE BASIN OF THE BALTIC SEA - UNECE
Content: Vistula. 194,424. Baltic Sea. BY, PL, SK, UA. - Bug. 39,400. Vistula. BY, PL, UA. - Dunajec. 4726.7. Vistula. PL, SK. -Poprad. 2,077. Dunajec. PL, SK. Oder. 118,861. Baltic Sea. CZ, DE, PL. - Neisse ... Oder. CZ, DE, PL. - Olse ... Oder. CZ, PL. 1 The assessment of water bodies in italics was not included in the present publication. 2 For the Venta River Basin District, which includes the basins of the Barta/Bartuva and Sventoji rivers. Oulu. Lulea. Rovaniemi. Lake. Oulujarvi. Lake. Tornetrask. Torne. Oulujoki.
]
Output: The query asks about drainage basin of ounasjoki. The most relevant knowledge entry that contain basin is "Venta River Basin District, which includes the basins of the Barta/Bartuva and Sventoji rivers."
From this result we know the drainage basin shall be Venta River Basin.
Therefore, the predicted answer is Venta River Basin.


Query: What is the typical diameter (in centimetre) of tennis?
Knowledge: [
Title: What Size Is A Tennis Ball In Cm? - Metro League
Content: To Recap. A tennis ball is typically about 2 cm in diameter. Similar Posts: What Is A Junk Ball In Tennis?
How tall is a tennis ball? Tennis Balls come in different sizes, some as small as 2.575"-2.7" (6.54-6.86 cm) and others up to 8 inches (20 cm). The mass of a tennis ball must be between 1.975-2.095 oz (56-59 g).

Title: Tennis Ball Dimensions & Drawings | Dimensions.com
Content: Tennis Balls have a diameter of 2.575"-2.7" (6.54-6.86 cm) and circumference of 8.09"-8.48" (20.6-21.5 cm). The mass of a Tennis Ball must be between 1.975-2.095 oz (56-59.4 g).
Tennis Balls have a diameter of 2.575"-2.7" (6.54-6.86 cm) and circumference of 8.09"-8.48" (20.6-21.5 cm). The mass of a Tennis Ball must be between 1.975-2.095 oz (56-59.4 g). A Tennis Ball is a ball designed for the sport of tennis.

Title: Tennis ball - Wikipedia
Content: Modern tennis balls must conform to certain criteria for size, weight, deformation, and bounce criteria to be approved for regulation play. The International Tennis Federation (ITF) defines the official diameter as 6.54-6.86 cm (2.57-2.70 inches). Balls must have masses in the range 56.0-59.4 g (1.98-2.10 ounces).
]
Output: The query asks about diameter of tennis (in centimetre). the most relevant knowledge entry about diameter of tennis is "tennis balls have a diameter of 2.575"-2.7" (6.54-6.86 cm) and circumference of 8.09"-8.48" (20.6-21.5 cm)".
As the query ask about centimetre, cm. From this result we know the diameter shall be 6.54 - 6.86.
Therefore, the predicted answer is 6.54 - 6.86.



Query: Who is the inventor of women netball, sport?
Knowledge: [
Title: 
Content: History of netball - Wikipedia

In 1893, Martina Bergman-osterberg informally introduced one version of basketball to her female physical training students at the Hampstead Physical Training College in London, after having seen the game being played in the United States.

Title: History of netball - Wikipedia
Content: In 1893, Martina Bergman-osterberg informally introduced one version of basketball to her female physical training students at the Hampstead Physical Training College in London, after having seen the game being played in the United States. Madame osterberg advocated physical fitness for women to better prepare them for motherhood and in the wider context of women's emancipation.

Title: Netball - Wikipedia
Content: A common misunderstanding of netball's origins has resulted in the mistaken belief that netball was created to prevent women from playing basketball. However, netball's development traces back to American sports teacher Clara Gregory Baer's misinterpretation of the basketball rule book in 1895.
History. Netball's early development emerged from Clara Baer's misinterpretation of the early rules of James Naismith's new sport of basketball (which he developed while studying in Massachusetts) and eventually evol

Title: History of Netball - World Netball
Content: Women's indoor basketball began exactly two days later when female teachers to the gym were captivated by the game but it wasn't until 1895 that the current game of netball was well and truly shaped. When Clara Baer, a sports teacher in New Orleans, wrote to Naismith asking for a copy of the rules, the subsequent rules package contained a drawing of
]
Output: The query asks about inventor of women netball. The most relevant knowledge entry about women netball inventor is "In 1893, Martina Bergman-Osterberg informally introduced one version of basketball to her female physical training students".
From the result, we know the inventor shall be Martina Bergman-Osterberg.
Therefore, the predicted answer is Martina Bergman-Osterberg.


Query: How many elevators does torre picasso have?
Knowledge: [
Title: 
Content: Torre Picasso | Turismo Madrid

The interior of the Picasso Tower houses offices designed as intelligent spaces equipped with the highest technology, comfort and use of space. It has 18 lifts, divided into three groups of six.

Title: Torre Picasso - Wikipedia
Content: 26 elevators; 18 serve office floors divided into three zones:

- 1st-18th floors at 2.5 m/s (8.20 ft/s)
- 18th-32nd floors at 4 m/s (13.12 ft/s)
- 32nd-43rd floors at 6 m/s (19.69 ft/s) (fastest in Spain)

Title: Torre Picasso - Field Trip
Content: 26 elevators, of which 18 to office floors in 3 groups of 6:

- 1st-18th floors at 2.5 m/s (8.20 ft/s)
- 18th-32nd floors at 4 m/s (13.12 ft/s)
- 32nd-43rd floors at 6 m/s (19.69 ft/s) (apparently fastest in Spain)

Title: Torre Picasso - Wikiwand
Content: The building as seen from the junction of the Paseo de la Castellana and the Plaza de Pablo Ruiz Picasso. 26 elevators; 18 serve office floors divided into three zones: 1st-18th floors at 2.5 m/s (8.20 ft/s) 18th-32nd floors at 4 m/s (13.12 ft/s)

]
Output: The query asks about number of elevators in torre picasso. the most relevant knowledge entry about number of elevators in torre picasso is "26 elevators; 18 serve office floors divided into three zones:".
From the result, we know the number of elevators shall be 26.
therefore, the predicted answer is 26.
"""
\end{lstlisting}

\begin{lstlisting}[language=Python,caption={\small{Workflow of AVIS (code snippets)
}},label={code:workflow},floatplacement=H]
class MemoryState:
  state: str = ''
  traversed_actions: list = []
  query: str = ''
  context: str = ''

  def __init__(self, state, query = '', context = ''):
    self.state = state
    self.query = query
    self.context = context

def plan(transition_graph, cur_memory, lens_res, retr_res):
  action_list = [a for a in transition_graph[cur_memory.state] if a not in cur_memory.traversed_actions]
  action_prompt = ''
  for a in action_list:
    action_prompt += '  --' + a + ': ' + task_instructions[a] + '\n'
  prompt_example = ""
  for a in action_list:
    prompt_example += action_prompt_dict[a] + "\n"
  action_prompt = planner_prompt % (cur_memory.query, action_prompt, prompt_example, cur_memory.query, cur_memory.context)
  action = api_utils.call_palm(action_prompt)[0]

  instruction = []
  if action in require_instruction:
    exclude_ids = cur_memory.traversed_actions:
    prompt = instruction_prompt(cur_memory.query, lens_res, exclude_ids)
    res = api_utils.call_palm(prompt)[0]
    reason = parse_reason('the query asks about ' + reason)
    instruction = [reason, res]
  return action, instruction

def avis_execution(d):
  state = 'START'
  answer = None

  prompt = question_decomposition_prompt + 'Question: ' + q + '\n'
  res = api_utils.call_palm(prompt)[0]

  vqi = res.find('Visual: ')
  kqi = res.find('Knowledge: ')
  vq = res[vqi + 8: kqi-1]
  kq = res[kqi+11:]
  
  working_memory = [MemoryState(state = 'START', query = vq, context = lens_res[0])]
  while not answer:
    cur_memory = working_memory[-1]
    action, instruction = plan(transition_graph, cur_memory, lens_res, retr_res)
    exec_res = execute(action, instruction, lens_res, retr_res)
    res = reason(exec_res)
    if 'not informative' in res:
      cur_memory.traversed_actions += [action]
    elif 'answer is' in res:
      answer = res[10:]
    else:
      working_memory += [MemoryState(state = action, query = kq, context = res)]
  return answer
\end{lstlisting}

\section{Error Analysis}~\label{sec:error}

\begin{figure}[h!]
    \centering
    \includegraphics[width=120mm]{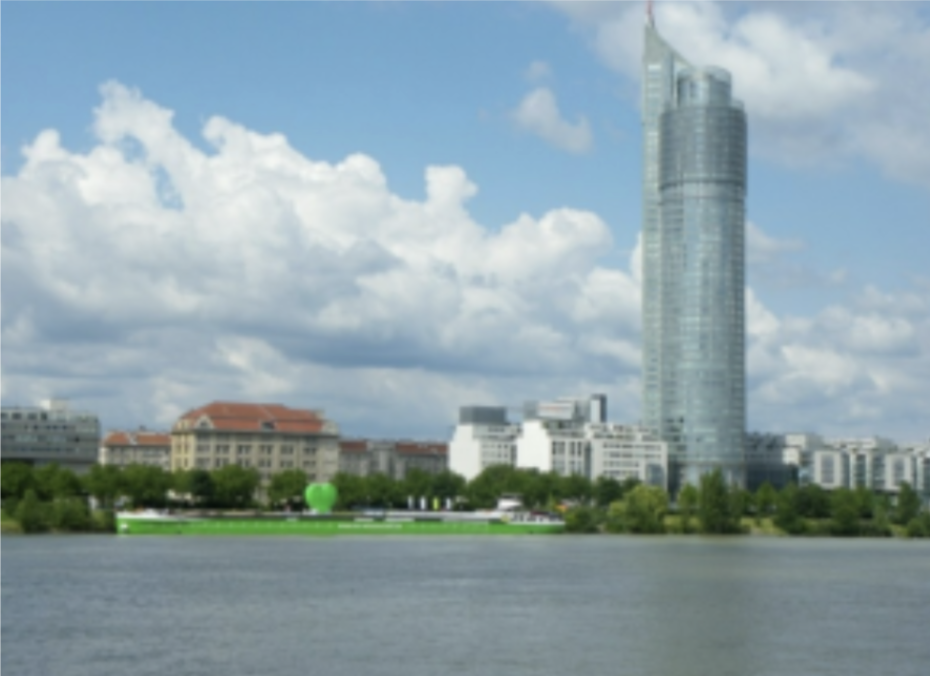}
    \caption{\textbf{Error Type}: LLM planning module miss important details. \textbf{Qustion}: What country does this building belong to? \textbf{Correct answer}: Austria; \textbf{Predicted answer}: United States}
    \label{fig:enter-label}
\end{figure}
\begin{itemize}
    \item  Step 1: object\_select, query = What is this building?
    \item Step 2: image\_search, query = what is this building?

    Retrievals: [ Millennium tower wien hi-res stock photography and images - Alamy (score=48.4), Millennium Tower Office Building in Vienna, Austria......

    Reason: The query asks about the building. From Object, there is a mention about "Millennium Tower" with high score. From Retrievals, there are many mentions about "Millennium Tower" with high score. Therefore, the predicted answer is Millennium Tower, building.

    \item Step3: Google Search, query = country of Millennium Tower, building?

    Title: Number 8: Millennium Tower in SoMa, San Francisco - San Francisco YIMBY Content: The eighth tallest skyscraper in the Bay Area planned or built is the Millennium Tower...

    From the search result, "The Millennium Tower is a skyscraper in San Francisco, California", the predicted answer is United States.

    yes, answerable, Output = United States
\end{itemize}

\begin{figure}[h!]
    \centering
    \includegraphics[width=80mm]{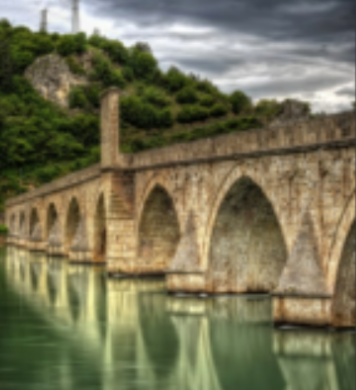}
    \caption{\textbf{Error Type}: LLM QA module extracts wrong evidence. \textbf{Qustion}: What is the architectural style of this bridge?  \textbf{Correct answer}: Ottoman architecture; \textbf{Predicted answer}: turkish monumental architecture}
    \label{fig:enter-label}
\end{figure}

\begin{itemize}
    \item image\_search query=what is this bridge?

    Retrievals: 
      high confidence: Na Drini ćuprija - Zadužbina Mehmed-paše Sokolovića - Furaj.ba | S nama u avanturu (score=87.2),
      high confidence: File:NKD279 Most Mehmed Paše Sokolovića Višegrad 5.jpg - Wikimedia Commons (score=91.3),
      ......]

    yes, answerable. The query asks about the bridge. From Retrievals, there are many mentionings about "Mehmed Paša Sokolović Bridge". Therefore, the predicted answer is Mehmed Paša Sokolović Bridge, bridge.

    \item Google Search, query=architectural style of Mehmed Paša Sokolović Bridge, bridge?

    Title: Mehmed Pasha Sokolovic Bridge – Višegrad, Bosnia and Herzegovina - Atlas Obscura
    Content: Celebrated example of Ottoman architecture that also served as a site for brutal massacres. Been Here? ......

    yes, answerable. Question asks about architectural style of Mehmed Paša Sokolović Bridge, bridge. From the search resutls, the most relevant knowledge entry about architectural style is "It is characteristic of the apogee of Turkish monumental architecture and civil engineering".
    Therefore, the predicted answer is Turkish monumental architecture.

\end{itemize}

\begin{figure}[h!]
    \centering
    \includegraphics[width=100mm]{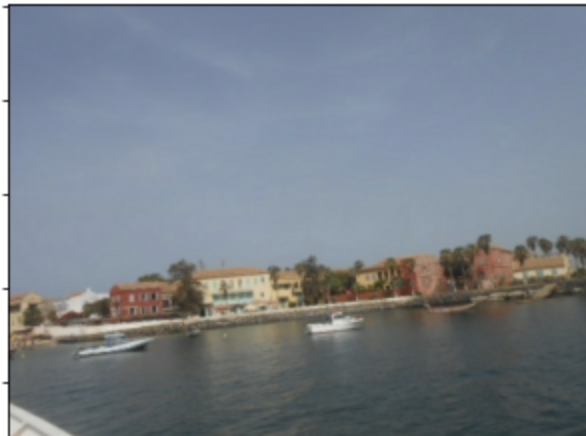}
    \caption{\textbf{Error Type}: Tool provides incorrect information. \textbf{Qustion}: Which body of water is this place located in or next to?  \textbf{Correct answer}: Atlantic; \textbf{Predicted answer}: land}
    \label{fig:enter-label}
\end{figure}

\begin{itemize}
    \item Step 1: identical\_image\_search, query = What is this place?
Retrievals: [
  high confidence: Reportage entre les murs de Mariama Bâ, internat d’excellence pour jeunes filles sénégalaises - Debout Congolaises (score=95.3),
]
not informative, \textbf{backtrack}
\item Step 2: object\_select, query = What is this place?
the query asks about the place. the image contains boats, water, trees and buildings. there is no explicit information about the place.
therefore, the predicted object \#id is 0, which is whole image.

Step 3: image\_search, query = What is this place?
Object: [
  a group of boats floating on top of a body of water . There are two white boats on the water. Behind that there are trees and buildings. (Caption, whole image)
]
Retrievals: [
  high confidence: Reportage entre les murs de Mariama Bâ, internat d’excellence pour jeunes filles sénégalaises - Debout Congolaises (score=95.3),
]
not informative, \textbf{backtrack}
\item Step 4: vqa, query = What is this place?
output=a lake
yes, answerable
\item Step 5: llm\_qa, query = body of water is a lake located in or next to?
output=land
\end{itemize}

\end{document}